\documentclass{article}

\usepackage[nonatbib, final]{neurips_2025_ml4ps}
\usepackage[utf8]{inputenc}     
\usepackage[T1]{fontenc}        
\usepackage{microtype}          
\usepackage{booktabs}           
\usepackage{graphicx}           
\usepackage{subcaption}         
\usepackage[inline]{enumitem}   
\usepackage{csquotes}           
\usepackage{amsmath}            
\usepackage{amsfonts}           
\usepackage{amssymb}            
\usepackage{amsthm}             
\usepackage{bm}                 
\usepackage{esint}              
\usepackage{nicefrac}           
\usepackage{siunitx}            
\usepackage[dvipsnames]{xcolor} 
\usepackage[colorlinks=true, allcolors=Blue]{hyperref} 
\usepackage{lipsum}             

\RequirePackage[ruled]{algorithm}
\RequirePackage[noend]{algpseudocode}

\makeatletter
\renewcommand\fs@ruled{
    \def\@fs@cfont{\bfseries}\let\@fs@capt\floatc@ruled%
    \def\@fs@pre{\hrule height \heavyrulewidth depth 0pt \kern 4pt}%
    \def\@fs@mid{\kern 4pt \hrule height \heavyrulewidth depth 0pt \kern 4pt}%
    \def\@fs@post{\kern 4pt \hrule height \heavyrulewidth depth 0pt \relax}%
    \let\@fs@iftopcapt\iftrue%
}
\makeatother

\algrenewcommand{\algorithmiccomment}[1]{\hfill #1}
\algrenewcommand{\alglinenumber}[1]{\footnotesize{#1}}

\usepackage[
    backend=biber,
    style=numeric-comp,
    sorting=none,
    maxcitenames=1,
    maxbibnames=2,
    doi=false,
    isbn=false,
    url=false,
]{biblatex}

\DeclareFieldFormat*{title}{\enquote{#1}}
\DeclareFieldFormat*{citetitle}{\enquote{#1}}

\newcommand*{\grad}[1]{\nabla_{\!{#1}}}

\title{Appa: Bending Weather Dynamics with Latent Diffusion Models for Global Data Assimilation}

\author{%
    Gérôme Andry \quad Sacha Lewin \quad François Rozet \quad \textbf{Omer Rochman} \quad \textbf{Victor Mangeleer} \\[1em] 
    \textbf{Matthias Pirlet} \quad \textbf{Elise Faulx} \quad \textbf{Marilaure Grégoire} \quad \textbf{Gilles Louppe} \\ \\ University of Liège
}

\begin{document}
\maketitle

\begin{abstract}
Deep learning has advanced weather forecasting~\cite{pathak2022fourcastnet, chen2023fuxi, lam2023learning, price2025probabilistic, lang2024aifs, bodnar2024foundation, nan2025langya, surya2024samudra}, but accurate predictions first require identifying the current state of the atmosphere from observational data. \changes{In this work, we} introduce Appa, a score-based data assimilation model generating global atmospheric trajectories at 0.25\si{\degree} resolution and 1-hour intervals. Powered by a 565M-parameter latent diffusion model trained on ERA5, Appa can be conditioned on arbitrary observations to infer \changes{plausible trajectories}, without retraining. Our probabilistic framework handles reanalysis, filtering, and forecasting, within a single model, producing physically consistent reconstructions from various inputs. Results establish latent score-based data assimilation as a promising foundation for future global atmospheric modeling systems.
\end{abstract}

\section{Introduction} \label{sec:introduction}
Data assimilation combines observational data with physical models to estimate atmospheric states. Formally, let $x^{1:L} = (x^1, x^2, \dots, x^L) \in \mathbb{R}^{L \times V \times C}$ denote a trajectory of $L$ atmospheric states, each represented as $C$ physical fields over a mesh of $V$ vertices. Let $p(x^1)$ be the initial state prior and $p(x^{i+1} \mid x^i)$ the transition dynamics. Observations $y \in \mathbb{R}^{M}$ of the state trajectory $x^{1:L}$ follow an observation process $p(y \mid x^{1:L})$, generally formulated as $y = \mathcal{M}(x^{1:L}) + \eta$, where the measurement function $\mathcal{M}: \mathbb{R}^{L \times V \times C} \mapsto \mathbb{R}^M$ might be non-linear and $\eta \in \mathbb{R}^M$ represents observational error that accounts for instrumental noise and systematic uncertainties. The goal of data assimilation is to infer plausible trajectories $x^{1:L}$ consistent with the observations, that is, to estimate the trajectory posterior
\begin{equation}
    p(x^{1:L} \mid y) = \frac{p(y \mid x^{1:L})}{p(y)} \, p(x^1) \prod_{i=1}^{L-1} p(x^{i+1} \mid x^i) \, .
    \label{eq:general-assim}
\end{equation}
While Eq.~\eqref{eq:general-assim} defines the posterior inference problem generically, assimilation tasks correspond to specific choices of \changes{states $x^i$} and observations $y$ \changes{to take into account}. In this work, we focus on three practical cases:
\begin{align}
    \textbf{Reanalysis:} \quad & x^{1:L} \sim p(x^{1:L} \mid y^{1:L}), \label{eq:reanalysis}\\
    \textbf{Filtering:} \quad & x^L \sim p(x^L \mid y^{1:L}), \label{eq:filtering}\\
    \textbf{Forecasting:} \quad & x^{K+1:L} \sim p(x^{K+1:L} \mid y^{1:K}) \;\;\text{or}\;\; p(x^{K+1:L} \mid x^{1:K}). \label{eq:forecast}
\end{align}
Reanalysis aims to reconstruct full trajectories from partial historical observations of the same time segment. \changes{The primary purpose of reanalysis is to create datasets of historical data for the land, atmospheres, and oceans. These datasets enable scientists to better monitor and understand the climate, conduct surveys, and develop new weather models.} Filtering, in contrast, only \changes{infers the posterior distribution of the} current state, obtained as a marginal of the reanalysis posterior. Forecasting, \changes{as its name suggests}, extends beyond the observed segment, producing posterior distributions over future states. \changes{The forecasting problem is often initialized by an external estimation of the current state $x^K$ or decomposed into first estimating this current state from $y^{1:K}$, and then predicting its evolution.}

\changes{Traditional methods like 4D-Var~\cite{lorenc1986analysis, dimet1986variational, tremolet2006accounting,tremolet2007modelerror,fisher2011weakconstraint} and ensemble Kalman filters~\cite{hunt2007efficient} are effective but rely on linearizations, require expensive differentiation, and provide point estimates rather than full posterior distributions~\cite{carrassi2018data}. Recent data-driven approaches~\cite{cheng2023machine, huang2024diffda, xu2024fuxida, fablet2021joint, andrychowicz2023deep} integrate deep learning into assimilation or forecast directly from observations, but suffer from limited resolution, lack of uncertainty quantification, and require retraining for new observation configurations.}

\section{Appa} \label{sec:appa}

Appa combines score-based data assimilation~\cite{rozet2023scorebased, rozet2023scorebaseda, schmidt2024sdanvidia} with latent diffusion models for physics emulation~\cite{rozet2025lost}, scaled to the \changes{global} atmospheric system at 0.25\si{\degree} resolution and 1-hour intervals, with 6 surface variables and 5 atmospheric variables across 13 pressure levels.

\paragraph{Architecture}
Appa consists of a 340M-parameter encoder-decoder pair $(E_\psi, D_\psi)$ that compresses atmospheric states $x^i$ by a factor $530$ into a latent representations \changes{$z^i \sim \mathcal{N}(z^i\mid E_\psi(x^i), \sigma_z^2I)$}. This reduces the dimensionality from \changes{$\mathcal{O}(10^8)$} elements per \changes{atmospheric} state (\changes{$\mathcal{O}(10^{10})$} for 4 days at 1-hour resolution) to \changes{$\mathcal{O}(10^5)$} elements per latent state (\changes{$\mathcal{O}(10^7)$} for 4 days), enabling efficient generation and inference in the latent space. The encoder-decoder pair is trained with a latitude- and level-weighted mean squared error loss~\cite{lam2023learning, price2025probabilistic}.

The autoencoder is paired with a 225M-parameter \changes{diffusion transformer (DiT)} \cite{peebles2023scalable} that operates on \changes{windows} of $W = 24$ consecutive latent states. 
Following~\cite{karras2022elucidating,rozet2023scorebaseda}, we train a denoiser to estimate the denoising posterior mean $\mathbb{E}\left[z^{i:i+W} \mid z^{i:i+W}_t\right]$ which, via Tweedie's first-order formula, provides the prior score function $\grad{z^{i:i+W}_t} \log p(z^{i:i+W}_t)$ needed for the reverse diffusion sampling process. For a variance exploding diffusion process~\cite{song2020scorebased}, we have
\begin{equation}\label{eq:tweedie1}
    \mathbb{E}\left[z^{i:i+W} \mid z^{i:i+W}_t\right] = z^{i:i+W}_t + \sigma^2_t\grad{z^{i:i+W}_t} \log p(z^{i:i+W}_t).
\end{equation}
\changes{As in the SDA \cite{rozet2023scorebased, rozet2023scorebaseda} framework, the score over trajectories $z^{i:i+L}_t$ is approximated by composing the scores over windows. We generalize
SDA's composition algorithm by introducing a time stride $\Delta \geq 1$ between consecutive
windows. Using a larger stride reduces the window overlap and, therefore, the number
of network evaluations. Detailed algorithms for training and composing local scores can be found in \ref{alg:sda}}.

\paragraph{Sampling conditionally on weather observations} 
To \changes{sample} from the posterior $p(z^{1:L} \mid y)$, we replace the prior score in the reverse diffusion sampling process with the posterior score
\begin{equation} \label{eq:score-bayes}
    \grad{z^{1:L}_t} \log p(z^{1:L}_t \mid y) = \grad{z^{1:L}_t} \log p(z^{1:L}_t) + \grad{z^{1:L}_t} \log p(y \mid z^{1:L}_t) \, .
\end{equation}
The \changes{prior score} $ \grad{z^{1:L}_t} \log p(z^{1:L}_t)$ \changes{is obtained} from the denoiser \changes{and combination of scores over windows}, while the \changes{likelihood score} $\grad{z^{1:L}_t} \log p(y \mid z^{1:L}_t)$ can be approximated without retraining \cite{chung2022diffusion, rozet2023scorebased, rozet2024learning, daras2024survey} under moderate assumptions on the observation process.

The key challenge is that the observation process $p(y \mid x^{1:L})$ is defined for atmospheric states $x^i$ rather than latent states $z^i$. For observation \changes{sequences} $y^{1:L}$ of the form $y^i = \mathcal{M}^i(x^i) + \eta^i$, we approximate the mapping from $z^i$ to $y^i$ as the composition of the decoder $D_\psi$ and measurement operator $\mathcal{M}^i$. Formally,
\begin{equation}\label{eq:latent-observation-process}
    p(y^{1:L} \mid z^{1:L}) \approx \mathcal{N}(y^{1:L} \mid \mathcal{A}(z^{1:L}), \Sigma_{y}) \, ,
\end{equation}
\changes{such that} $\mathcal{A}(z^{1:L}) = (\mathcal{M}^1(D_\psi(z^1)) \cdots \mathcal{M}^L(D_\psi(z^L)))^\top$ and $\Sigma_y$ is the covariance of $\eta^{1:L}$. 
\changes{With this formulation}, off-the-shelf posterior sampling algorithms~\cite{daras2024survey} can be used for generating atmospheric trajectories conditionally on observational data. In this work, we adapt moment matching posterior sampling (MMPS), originally proposed by \textcite{rozet2024learning} for linear observation operators. In our case, since $\mathcal{A}$ is non-linear, we use its Jacobian $A$ in the estimate of the covariance, yielding the approximation of the perturbed likelihood
\begin{equation} \label{eq:likelihood-mmps}
    p(y^{1:L} \mid z^{1:L}_t) \approx \mathcal{N}(y \mid \mathcal{A}(\mathbb{E}[z^{1:L} \mid z^{1:L}_t]), \Sigma_y + A \mathbb{V}[z^{1:L} \mid z^{1:L}_t] A^\top) \, .
\end{equation}

\changes{After inference in the latent space, the generated trajectories $z^{1:L}$ are decoded back to the atmospheric space via the decoder $\hat{x}_i = D_\psi(z_i)$.}

\section{Experiments} \label{sec:experiments}
We train and evaluate Appa on the ERA5 reanalysis \changes{dataset} \cite{hersbach2018era5} following standard chronological splitting: 1993--2021 for training, 2020 for validation, and 2021 for testing. Below, we evaluate the latent representation quality, assimilation performance across reanalysis, filtering and forecasting tasks, and compare against existing methods.

\begin{figure}
    \centering
    \includegraphics[width=\linewidth]{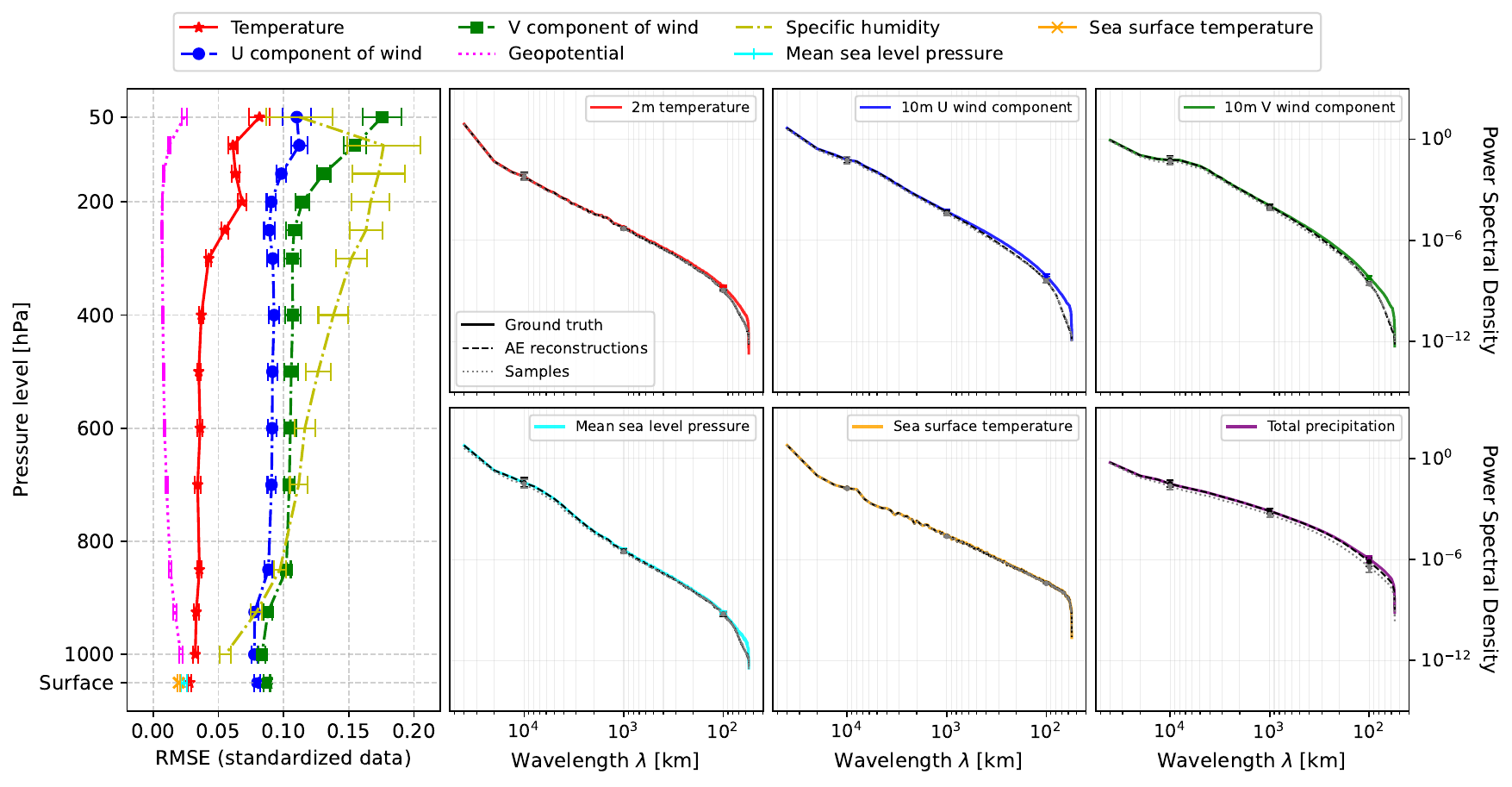}
    \caption{\changes{(Left) Standardized autoencoder reconstruction RMSEs. Lower-frequency fields (temperature, geopotential) are reconstructed more accurately than volatile fields (humidity, winds). Near-surface fields benefit from altitude-weighting. (Right) Power spectral density comparison of ground truth, autoencoder reconstructions, and samples generated from Appa’s prior. Median and percentile ranges show close alignment across scales, with deviations below 100 km (3 to 4 grid cells).}}
    \label{fig:ae-rmse}
\end{figure}

\paragraph{Latent representation}
Despite \changes{the} $530 \times$ compression factor, \changes{standardized} reconstruction RMSEs are mostly below $0.1$, \changes{with slightly higher values for humidity and
winds, and lower ones for surface and low-altitude fields, as shown in Figure~\ref{fig:ae-rmse}}. \changes{Reconstructed} power spectra match ground truth closely, except at scales below 100 km where atmospheric energy is minimal. Compared to prior neural compression methods~\cite{mirowski2024neural}, our autoencoder achieves comparable performance.

\paragraph{Assimilation} 
We evaluate Appa across four scenarios: reanalysis, filtering, observational forecasting, and full-state forecasting. For the first three tasks, we assimilate both synthetic ground-station observations of all 6 surface variables and simulated scans of the 5 atmospheric variables across 13 pressure levels. The ground station network consists of 11,000 real-world measurement locations~\cite{gsod1999} covering roughly $1\%$ of grid points. Ground stations are sparse and globally distributed, while satellite orbital paths provide dense spatial coverage with restricted temporal and spatial reach. Observations are modeled as Gaussian distributions centered on the ERA5 ground truth, with \changes{noise levels} of $1\%$ for ground stations and $10\%$ for satellite measurements.

\begin{figure}
    \centering
    \includegraphics[width=\linewidth]{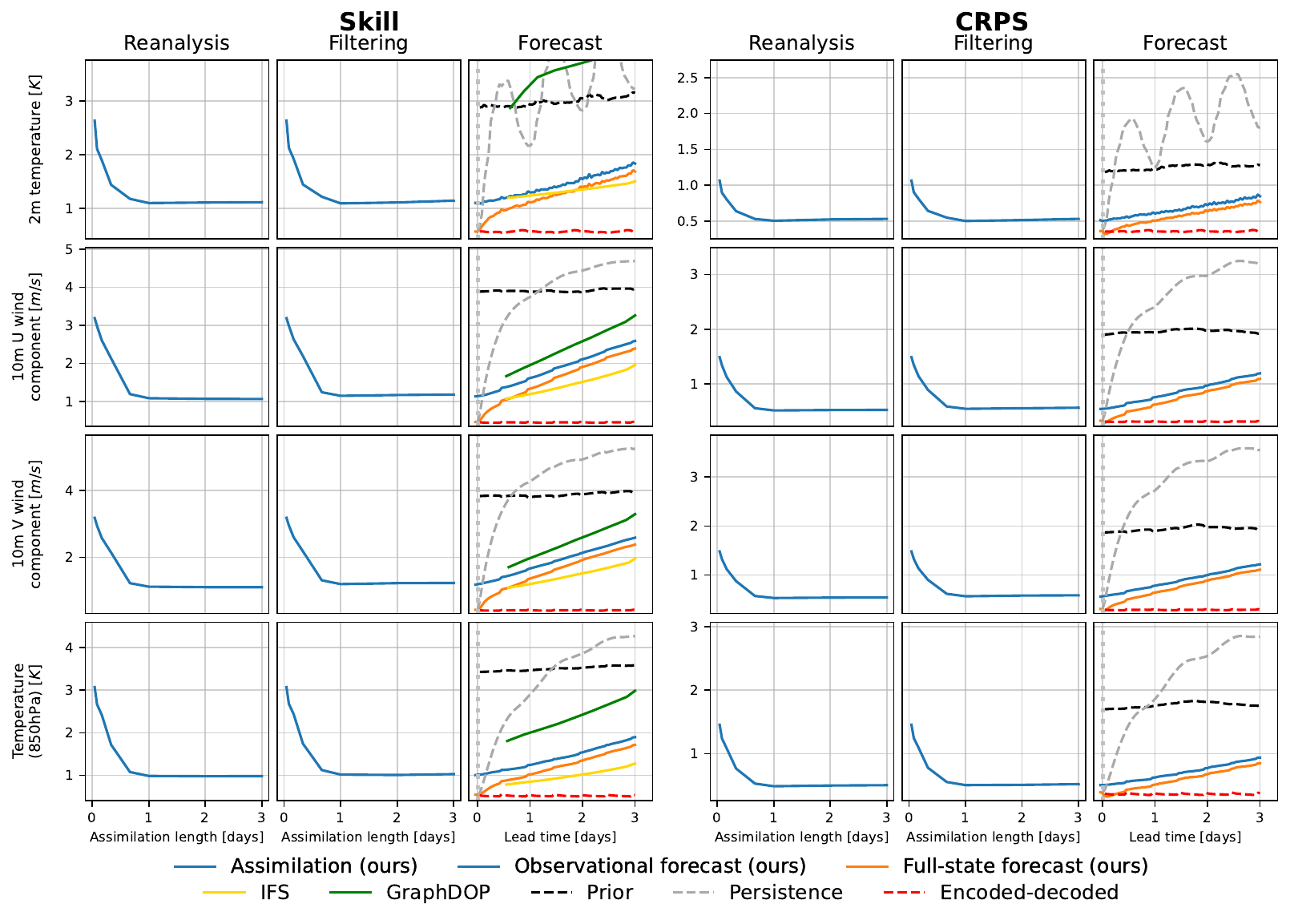}
    \caption{Skill and Continuous Ranked Probability Score (CRPS) for representative variables in January 2023. Detailed experimental setup can be found in \ref{app:forecast-details}. Reanalysis scores are averaged over assimilation windows \changes{$x^{1:L}$}, while filtering reports the last reanalyzed state. Both improve with longer \changes{assimilation windows, but eventually stagnate}. Forecasts gradually lose skill over lead time but remain superior to persistence and unconditional baselines. \changes{The initial skill level of the full-state forecast matches the compression error level.} IFS \cite{ifs} and GraphDOP~\cite{alexe2024graphdop} are shown for reference.}
    \label{fig:reanalysis-quant}
\end{figure}

Figure~\ref{fig:reanalysis-quant} summarizes Appa’s performance and further qualitative results can be found in \ref{app:gallery}. \changes{For reanalysis and filtering,} conditioning on longer assimilation windows improves both skill and CRPS but gains saturate beyond 24 hours. \changes{Forecasting's} skill decays gradually with lead time but remains significantly stronger than \changes{the persistence baseline}. Observational forecasts, conditioned on the last 12 hours of a day-long assimilation, \changes{start} at skill and CRPS levels comparable to the reanalysis plateau. Full-state forecasts, initialized from two complete states, start lower but eventually converge to similar performance over time. As expected, the initial skill for full-state forecasts is close to autoencoder reconstruction error levels. Overall, these results demonstrate that Appa successfully handles all assimilation and forecasting tasks within a unified probabilistic framework.

\changes{For short lead times, our forecasts reach skill levels comparable to IFS~\cite{ifs} while performing better than GraphDOP~\cite{alexe2024graphdop}. Notably, the forecast error growth lies between the two baselines: slightly steeper than IFS but below GraphDOP and closely tracking its slope. This suggests that compression does not introduce strong dynamical artifacts that impede the learning of the dynamics, even under different temporal setups (1-hour for Appa vs. 12-hour resolution for IFS) which corroborates with findings of \textcite{rozet2025lost}. While these results remain preliminary, they provide promising evidence that Appa can capture atmospheric dynamics at a level in between purely physical and purely data-driven approaches.}

\section{Discussion} \label{sec:discussion}
\paragraph{Summary} 
We introduce Appa, a latent score-based data assimilation framework that produces global atmospheric trajectories by operating in a compressed latent space. Appa can be conditioned on various types of observations without retraining, providing access to the full posterior distribution of \changes{consistent} trajectories. Our results show that Appa flexibly handles reanalysis, filtering, and forecasting within a single framework, producing competitive performance across scenarios without task-specific training or architectural modifications.

\paragraph{Limitations and future work}
While Appa demonstrates strong assimilation and forecasting capabilities, it remains a proof of concept and further improvements are needed to make it operational.
First, we should consider moving from simplified synthetic observations to realistic measurement, \changes{such as} satellite radiances. Improving physical \changes{consistency} is also critical, as compression inevitably degrades fine-scale information. Strategies such as localized assimilation or refined conditioning mechanisms may help. \changes{In terms of statistical assessment, the calibration of posterior distributions deserves further validation. Indeed, the approximations present in our method, notably while estimating the prior and likelihood scores, introduce errors which are complex to quantify.} The computational efficiency of \changes{observational tasks} remains a challenge as well, since conditioning requires repeated decoding steps, projecting observations directly into latent space could mitigate this bottleneck. Finally, our comparison to other models is still preliminary. Baselines for assimilation remain scarce, and fair evaluation \changes{against} IFS~\cite{ifs}, GraphDOP~\cite{alexe2024graphdop} and other models~\cite{price2025probabilistic, lam2023learning, raphaeli_silo_2025} \changes{would help position Appa within the spectrum of global atmospheric models.}

\begin{ack}
Gérôme Andry, François Rozet, Sacha Lewin, and Elise Faulx are research fellows of the {\it National Fund for Scientific Research} (F.R.S.-FNRS) and acknowledge its financial support. Omer Rochman gratefully acknowledges the financial support of the {\it Walloon Region} under Grant No. 2010235 (ARIAC by Digital Wallonia 4.AI). Victor Mangeleer is a research fellow part of the {\it Multiple Threats on Ocean Health} (MITHO) project and gratefully acknowledges funding from the {\it European Space Agency} (ESA).

The present research benefited from computational resources made available on Lucia, the Tier-1 supercomputer of the Walloon Region, infrastructure funded by the Walloon Region under the grant agreement n°1910247.
\end{ack}

\newpage
\section*{References}

\printbibliography[heading=none]

\newpage
\appendix

\section{Data} \label{app:data}

\subsection{ERA5} \label{app:data-era5}

ERA5 is a global deterministic reanalysis dataset from ECMWF that provides high-resolution (0.25°) hourly estimates of atmospheric, land, and oceanic variables from 1959 onward~\cite{hersbach2018era5}. It assimilates observations into a numerical weather prediction model using 4D-Var data assimilation.

For this work, we use a subset of ERA5 data, defined on a 0.25° equiangular grid with 13 pressure levels: 50, 100, 150, 200, 250, 300, 400, 500, 600, 700, 850, 925, and 1000 hPa. Due to storage limitations, we restrict the temporal coverage of the dataset to the 1993--2021 period, with data split into training (1993--2019), validation (2020), and testing (2021).

Table~\ref{tab:data-era5-variables} lists the selected variables. Some serve as both input and predicted features, while others provide contextual information (only input). Context variables are not predicted but help define the temporal and dynamic conditions under which predictions are made, improving model performance.

\begin{table*}[h]
\centering
\renewcommand{\arraystretch}{1.4} 
\setlength{\tabcolsep}{27pt} 
\caption{Input variables.} 
\label{tab:data-era5-variables}
\begin{tabular}{ccc}
\toprule
\textbf{Type} & \textbf{Variable Name} & \textbf{Role} \\
\midrule
Atmospheric & Temperature & Input/Predicted \\
Atmospheric & U-Wind Component & Input/Predicted \\
Atmospheric & V-Wind Component & Input/Predicted \\
Atmospheric & Geopotential & Input/Predicted \\
Atmospheric & Specific Humidity & Input/Predicted \\
Single & 2m Temperature & Input/Predicted \\
Single & 10m U-Wind Component & Input/Predicted \\
Single & 10m V-Wind Component & Input/Predicted \\
Single & Mean Sea Level Pressure & Input/Predicted \\
Single & Sea Surface Temperature & Input/Predicted \\
Single & Total Precipitation & Input/Predicted \\
\midrule
Clock & Local time of day & Input (Diffusion)\\
Clock & Elapsed year progress & Input (Diffusion)\\
\bottomrule
\end{tabular}
\end{table*}

\subsection{Data pre-processing} 

\paragraph{Standardization} Although the dynamics across the atmospheric column are correlated, each pressure level exhibits distinct statistical behavior. Thus, we computed the mean and standard deviation separately for each variable and at each pressure level, on the whole training dataset. We used these statistics to standardize our entire dataset and to rescale the output of Appa.

\paragraph{Handling missing values}
Since Sea Surface Temperature is undefined over land (NaN values), we replace these with zeros as a neutral placeholder after standardization.

\paragraph{Data availability} 
ERA5 data was downloaded from the WeatherBench2 platform, where Google has made them publicly available via Google Cloud Storage.

\newpage

\section{Technical details}

This section provides further technical details for training and inference. Our code will be made available with full reproducibility steps for both training and evaluation.

\subsection{Architectures}
We adapted architectures from \textcite{rozet2025lost} for both autoencoder and latent diffusion model. \changes{The encoder and decoder are fully convolutional neural networks and the diffusion
model is adapted from a diffusion transformer (DiT) \cite{peebles2023scalable}}.
\paragraph{Autoencoder}
The autoencoder compresses atmospheric states from the high-dimensional N320 grid ($721\times1440$ pixels with 71 channels) into a compact latent space ($23\times47$ pixels with 128 channels) via progressive downsampling and channel expansion in a fully convolutional architecture. To accommodate any spatial compression factor, the input is padded to the nearest compatible grid size. We apply periodic padding along longitude to respect global wrap-around, and constant zero padding along latitude to handle polar boundaries. \changes{The encoder-decoder pair is trained with a latitude-level-weighted mean squared error loss, following~\textcite{lam2023learning}.}

\paragraph{Latent denoiser}
The denoiser is a DiT that operates on 24 consecutive latent states. We first patch the latent sequence by a factor of 2 along the temporal axis, then flatten the spatial dimensions, yielding $23 \times 47 \times \nicefrac{24}{2} = 12{,}972$ tokens, each with 256 channels, which are passed to the DiT.

\begin{table}[h!]
    \centering
    \caption{Autoencoder training configuration}
    \begin{tabular*}{\textwidth}{p{3cm}l@{}}
        \toprule
        \textbf{Parameter} & \textbf{Value} \\
        \midrule
        Loss function & Latitude- and level-weighted mean squared error  \\
        Latent noise & $\sigma=0.01$ for regularization \\
        Optimizer & SOAP with initial learning rate $3\times 10^{-5}$ and linear decay\\
        Batch size & 64 samples per step \\
        Training duration & 95000 update steps (approximately 2 days) \\
        Hardware & 64× NVIDIA A100 40GB GPUs \\
        \bottomrule
    \end{tabular*}
    \label{tab:training-details}
\end{table}

\begin{table}[h!]
    \centering
    \caption{Denoiser training configuration}
    \begin{tabular*}{\textwidth}{p{3cm}l@{}}
        \toprule
        \textbf{Parameter} & \textbf{Value} \\
        \midrule
        Loss & Denoising score matching with rectified noise schedule \\
        Noise range & $\sigma_\text{min}=0.001$, $\sigma_\text{max}=1000$ \\
        Optimizer & Adam with initial learning rate $1 \times 10^{-4}$\\
        Batch size & 256 samples per step\\
        Training duration & 125000 update steps (approximately 5 days) \\
        Hardware & 64× NVIDIA A100 40GB GPUs \\
        \bottomrule
    \end{tabular*}
    \label{tab:denoiser-training}
\end{table}

\newpage

\subsection{Generating trajectories}\label{alg:sda}
\changes{
\begin{figure}[h!]
\begin{minipage}[t]{0.49\textwidth}%
\vspace{-1em}
\begin{algorithm}[H]
    \caption{Training $d_\phi(z^{i:i+W}_t)$} \label{alg:appa-training}
    \begin{algorithmic}[1]
        \For{$n = 1$ to $N$}
            \State $x^{1:L} \sim p(x^{1:L})$
            \State $i \sim \mathcal{U}(\{1, \dots, L-W\})$
            \State $t \sim \mathcal{U}(0, 1) \, , \varepsilon \sim \mathcal{N}(0, I)$
            \For{$j = i$ to $i + W$}
                \State $z_j \gets E_\psi(x_j)$
            \EndFor{}
            \State $z^{i:i+W}_t \gets \, z^{i:i+W} + \sigma_t \, \varepsilon$
            \State $\displaystyle \ell \gets \big\| d_\phi(z^{i:i+W}_t) - z^{i:i+W} \big\|^2_2$
            \State $\phi \gets \Call{SGD}{\phi, \grad{\phi} \, \ell}$
        \EndFor{}
	\end{algorithmic}
\end{algorithm}
\end{minipage}%
\hfill%
\begin{minipage}[t]{0.49\textwidth}%
\vspace{-1em}
\begin{algorithm}[H]
    \caption{Composing $d_\phi(z^{i:i+W}_t)$} \label{alg:appa-composing}
    \begin{algorithmic}[1]
        \Function{$d_\phi(z^{1:L}_t)$}{}
            \State $a \gets \nicefrac{(W - \Delta)}{2}$
            \State $b \gets a + \Delta$
            \State $E_{1:a} \gets d_\phi(z^{1:1+W}_t)[:\!a]$
            \For{$n = 0$ to $\nicefrac{(L - W)}{\Delta} + 1$}
                \State $i \gets 1 + n \Delta$
                \State $E_{i+a:i+b} \gets d_\phi(z^{i:i+W}_t)[a\!:\!b]$
            \EndFor{}
            \State $E_{L-W+b:L} \gets d_\phi(z^{L-W:L}_t)[b\!:]$
            \State \Return{$E_{1:L}$}
        \EndFunction{}
	\end{algorithmic}
\end{algorithm}
\end{minipage}
\end{figure}
}

\subsection{Assimilation tasks}\label{app:forecast-details}

\paragraph{Forecasting} Appa is trained to generate state windows of a given size, we use 24 hours. We first split the total window in two, the first part being the condition, and the second the part to be generated. Then, we use either the last states of a fully assimilated window (observational forecasting) or full encoded latent states (full-state forecasting). At each autoregressive steps, we use a sliding window to move $n$ steps forward after these steps were predicted. This mechanism offers a balance between conditioning window size and generation speed, as the former can be extended to provide more context but more limited speed (more generations required) or less context but faster total generation.

\paragraph{\changes{Evaluation setup}}
On Figure~\ref{fig:reanalysis-quant}, we report IFS \cite{ifs} and GraphDOP~\cite{alexe2024graphdop} skills using data from \textcite{alexe2024graphdop} as baselines. However a direct and fair comparison is difficult due to lack of experimental setup details. \textcite{alexe2024graphdop} mention that skills are computed over January 2023 forecasts for 6 different variables. To match the time period, we report the performance of our method over a 10-member ensemble of 3-day forecasts. We selected the first 8 days of January 2023 at midnight as starting timestamps. This makes a total of 8 dates with 10 members for reported metrics.

\newpage
\section{Evaluation metrics}

We follow conventional metrics computation for assimilation and forecasting performance. For a fair comparison with the literature, evaluation is performed using WeatherBench2 \cite{weatherbench2}. For assimilation, we average performance over the time steps.

\subsection{Skill}
Skill is computed as the root mean square error of the posterior mean of an ensemble compared to the ground-truth trajectory. For $K$ ensembles each consisting of $M$ predicted states $\hat{x}$ of resolution $H \times W$, ground truth $x$, the skill of a single time step is computed as
\[
    \text{Skill} = \sqrt{\dfrac{1}{KHW} \sum_{k=1}^{K} \sum_{i=1}^{H} \sum_{j=1}^{W} \left(x_{i,j}^k - \dfrac{1}{M}\sum_{m=1}^M \hat{x}_{i,j}^{k,m}\right)^2}.
\]

\subsection{Spread}
Ensemble spread is computed as the square root of the ensemble variance \cite{fortin2014should}:
\[
    \text{Spread} = \sqrt{\dfrac{1}{KHW} \sum_{k=1}^{K} \sum_{i=1}^{H} \sum_{j=1}^{W} \dfrac{1}{M-1} \sum_{m=1}^{M}\left(\hat{x}_{i,j}^{k,m} - \dfrac{1}{M}\sum_{n=1}^M \hat{x}_{i,j}^{k,n}\right)^2}.
\]

\subsection{Spread-skill ratio}
A well-calibrated forecast should have a (corrected for ensemble size) spread-skill ratio of 1, which is a necessary but not sufficient condition. Ratios below one indicate overconfident estimations. The correct ratio is defined as
\[
    \text{Ratio} = \sqrt{\dfrac{M+1}{M}} \dfrac{\text{Spread}}{\text{Skill}}.
\]

\subsection{Continuous ranked probability score (CRPS)}
The CRPS \cite{crps} is defined as
\[
    \text{CRPS} = \dfrac{1}{K} \sum_{k=1}^K \left(\dfrac{1}{M}\sum_{m=1}^M || \hat{x}^{k,m} - x^k ||_{L_1} - \dfrac{1}{2M(M-1)} \sum_{m=1}^M\sum_{n=1}^{M} || \hat{x}^{k,m} - \hat{x}^{k,n} ||_{L_1}\right).
\]
The first term penalizes the average divergence from the ground truth while the second term encourages spread. Therefore, the CRPS is lowest when the distribution of the ensemble matches the ground-truth distribution. Note the $L_1$ norm used, which means that in the deterministic case, this reduces to the MAE.

\vfill

\newpage

\section{Additional results}

\subsection{Power spectral density}

\begin{figure}[H]
    \centering
    \includegraphics[width=\linewidth]{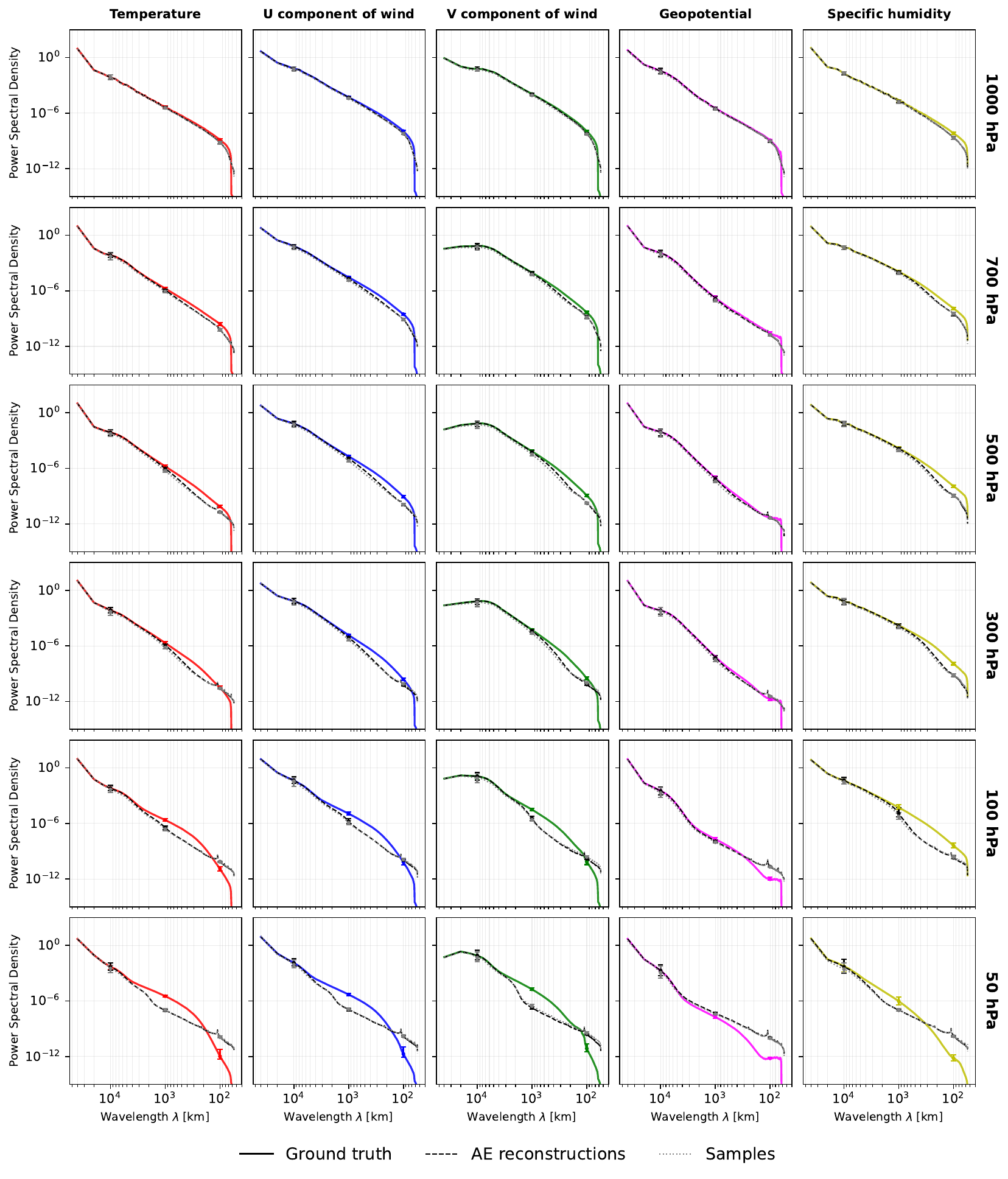}
    \caption{Power spectral density across wavelengths for atmospheric variables at selected pressure levels. Lines show median values and error bars indicate the 5th to 95th percentiles. The close alignment between the curves demonstrates that both the autoencoder and the diffusion model preserve the energy distribution across most spatial scales. Deviations begin to appear at wavelengths around 1000km, which corresponds to roughly 40 grid cells at our 0.25-degree resolution at the equator. These differences become more pronounced at smaller scales, suggesting that while large-scale atmospheric patterns are well-preserved, features spanning fewer than 40 grid cells show some energy loss in the compression and generation processes. Deviations become more pronounced at lower pressure levels, as the model prioritizes surface and low-altitude variables.}
\end{figure}

\begin{figure}[h!]
    \centering
    \includegraphics[width=\linewidth]{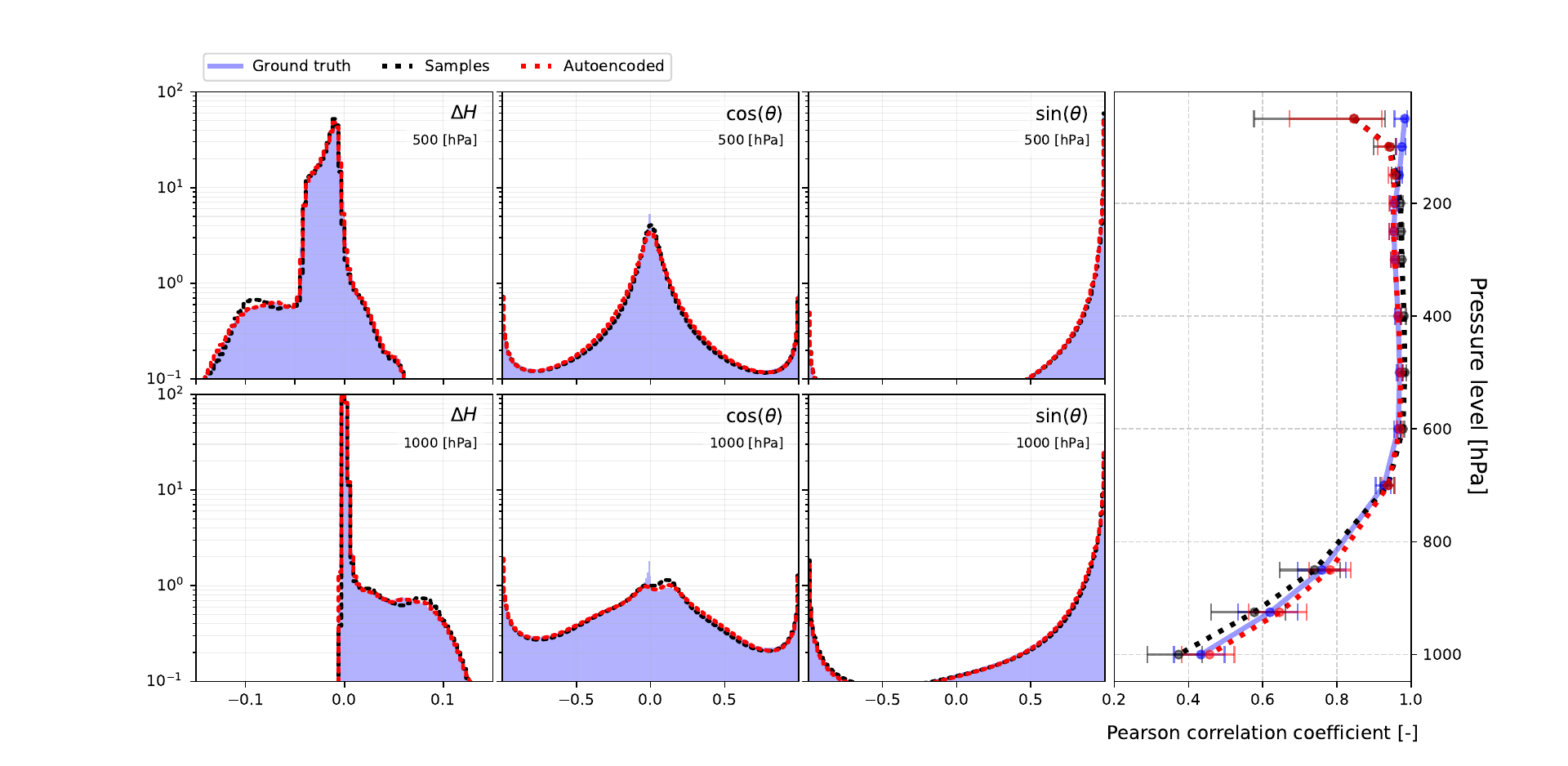}
    \caption{Physical consistency analysis of generated atmospheric states.
    (Top row) Analysis of altitude consistency at 500 hPa showing the difference $\Delta H$ between two independent altitude estimators, and geostrophic balance assessment through the cosine and sine of the angle $\theta$ between wind direction and geopotential gradients, demonstrating angles concentrated around 90°.
    (Bottom row) Same metrics at 1000 hPa demonstrating the presence of a significant ageostrophic component near the surface.
    (Right) Correlation coefficient between wind magnitude and geopotential gradient magnitude across pressure levels, showing strong correlation at upper levels (near 1) with a consistent decrease toward the surface in both ERA5 data (blue) and generated samples (black dots), confirming Appa's ability to capture complex physical relationships.}
    \label{fig:physical-consistency}
\end{figure}

\changes{\subsection{Physical consistency} \label{app:physical}}
To further evaluate physical consistency, we examine whether our model preserves important physical relationships between variables. First, we analyze the consistency between two different estimators of altitude at given pressure levels. Using the geopotential $\Phi$, altitude can be derived as 
\begin{equation}
    H = \frac{\Phi R_e}{g_{0}R_e-\Phi}
\end{equation} 
where $R_e$ is Earth's radius and $g_{0}$ is the Earth gravitational acceleration at the surface. Alternatively, the equation below (which relies on the ideal gas law and hydrostatic equation) relates altitude to pressure and temperature as 
\begin{equation} 
    \log{\frac{p_0}{p_H}} = \frac{M g_0}{R}\int_0^H \frac{1}{T_h} \partial{h},
\end{equation} 
where $R$ is the universal gas constant, $M$ is an approximation of the atmosphere's molar mass, $p_h, T_h$ are pressure and temperature at height $h$, and $p_0$, $T_0$ are the theoretical pressure and temperature at sea level. This integral can be approximated to extract $H$ using several assumptions about the temperature profile. When comparing these two estimators, Figure~\ref{fig:physical-consistency} shows that our generated samples maintain the same systematic differences $\Delta H$ as seen in ground-truth data. This remarkable consistency indicates that our model successfully preserves this physical relationships between temperature, pressure, and geopotential, allowing altitude to be estimated through two independent methods with nearly identical accuracy to the original ERA5 data.

Second, we examine the geostrophic balance
\begin{align*}
\label{eq:partialXY}
    \frac{\partial \Phi}{\partial x}&= \frac{4 \pi \Omega R_e}{N_x} \sin{\phi} \cos{\phi} u_g\\
    \frac{\partial \Phi}{\partial y}&= -\frac{2 \pi \Omega R_e}{N_y} \sin{\phi} v_g
\end{align*}
which is the theoretical equilibrium between pressure gradient forces and Coriolis forces that governs large-scale atmospheric motion. In the above system, $\phi$ is the latitude, $N_x$ and $N_y$ are the number of pixels along longitude and latitude, $\Omega$ is the magnitude of the Earth’s angular velocity, and $u_g$ and $v_g$ denote the eostrophic components of the wind. In this balance, in the absence of vertical motion, friction, and isobaric curvature, wind direction should be perpendicular to geopotential gradients, with wind speed proportional to gradient magnitude. This relationship can be expressed by comparing two quantities: (1) the angle $\theta$ between wind and geopotential gradients, which should approach 90° in geostrophic conditions, and (2) the correlation between wind speed magnitude and geopotential gradient magnitude, which should approach 1 in perfect geostrophic balance. Figure~\ref{fig:physical-consistency} shows that our generated samples accurately reproduce both aspects of this relationship. At 500 hPa, the approximate level of non-divergence with minimal surface friction effects, both ERA5 data and our generated samples show angles concentrated around 90°. Near the surface at 1000 hPa, where additional forces become significant, both datasets show a systematic deviation in angle. Similarly, the correlation between wind speed and geopotential gradient magnitudes in our samples closely matches the patterns observed in ERA5 data, exhibiting imperfect correlation only at lower pressure levels (explained by the level-weighted training) and following the same decreasing trend as pressure increases toward the surface, where ageostrophic components become more prominent.

These results demonstrate that our latent diffusion model not only preserves the statistical properties of atmospheric fields but also maintains important physical relationships between variables producing trajectories that are physically consistent and realistic. While these analyses confirm strong spatial consistency and physical fidelity, future work should extend our evaluation to more thoroughly quantify the temporal consistency of generated trajectories.

\vfill
\newpage

\subsection{Quantitative evaluation}
\changes{\subsubsection{Compression error}}
\begin{figure}[h!]
    \centering
    \includegraphics[width=\linewidth]{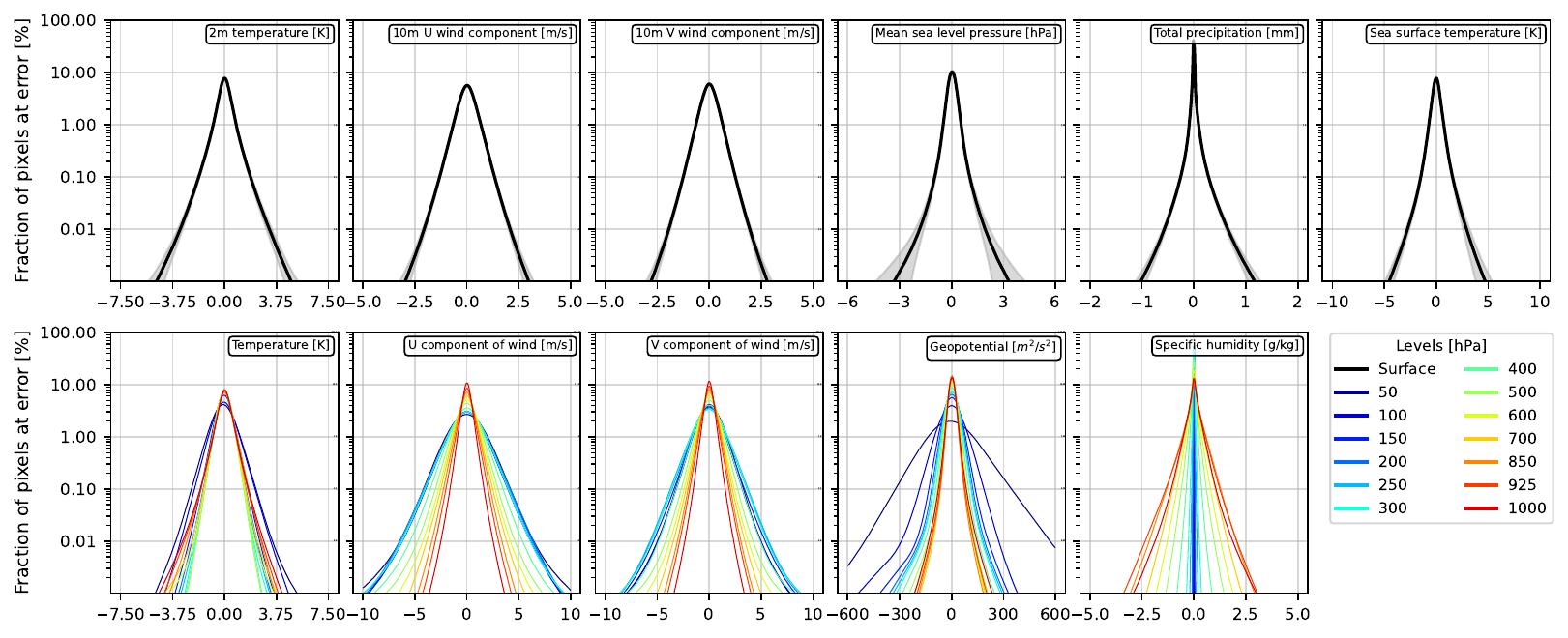}
    \caption{Signed reconstruction errors for surface variables (top) across all grid points and atmospheric variables (bottom) across all pressure levels. Shaded gray area for surface variables corresponds to error spread. In both cases, the concentrated distributions centered around zero demonstrate unbiased and precise predictions.  Given $721 \times 1440 = 1,038,240$ grid points, a $0.01 \%$ fraction on the y-axis corresponds to approximately 100 grid points, indicating that large errors are rare.}
    \label{fig:ae-signed-errors}
\end{figure}

\newpage 

\subsubsection{Additional variables}

\begin{figure}[h!]
    \centering
    \includegraphics[width=\linewidth]{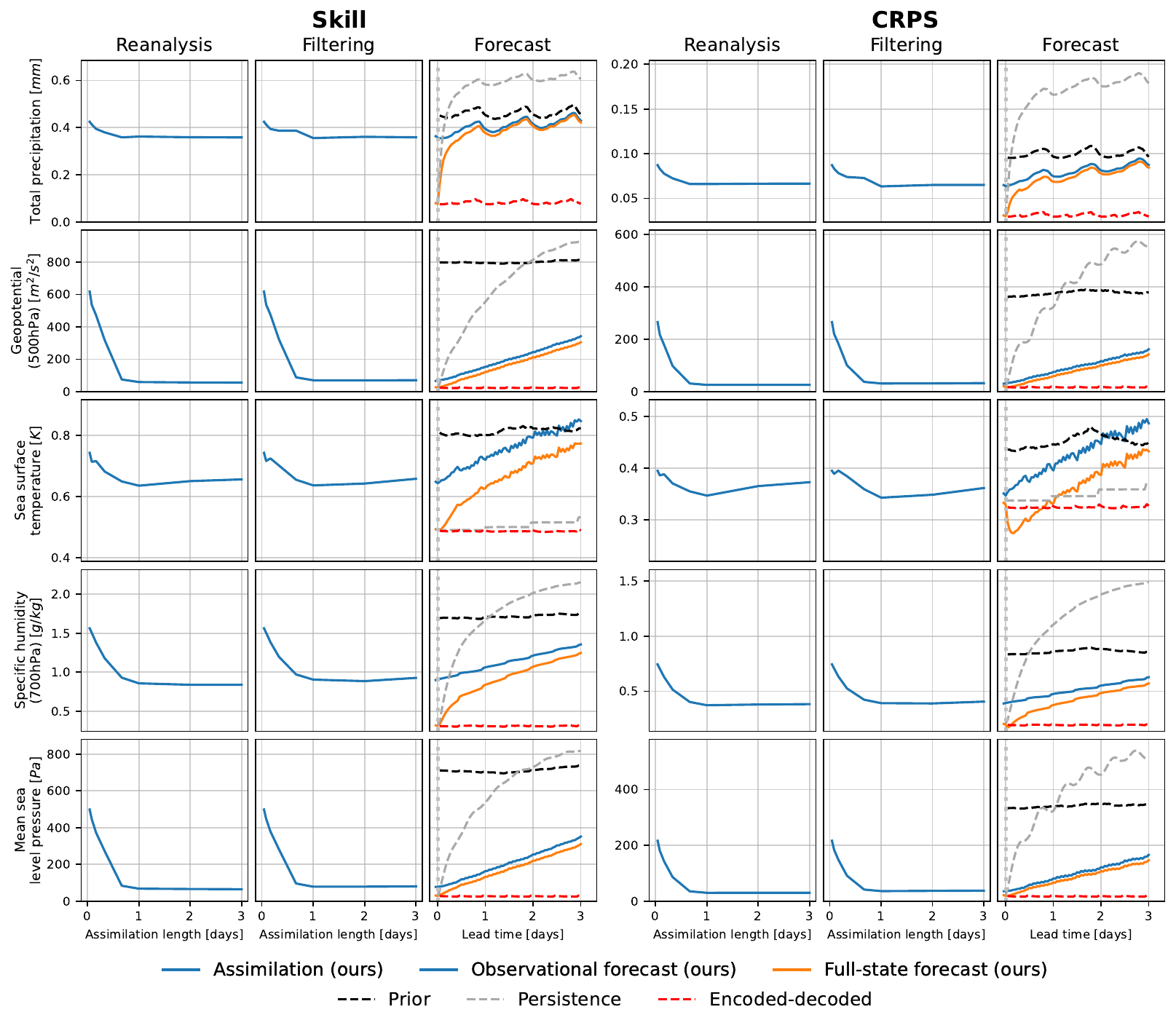}

    \caption{Quantitative evaluation of reanalysis, filtering, and forecasting for additional representative variables. (Left) Skill score and (Right) Continuous Ranked Probability Score (CRPS). Reanalysis scores are averaged over assimilation windows, while filtering reports the last reanalyzed state. Both improve with longer windows. Forecasts gradually lose skill over lead time but remain above persistence and unconditional baselines.}
\end{figure}

\newpage
\subsubsection{Spread and spread-skill ratios}

\begin{figure}[H]
    \centering
    \includegraphics[width=0.995\linewidth]{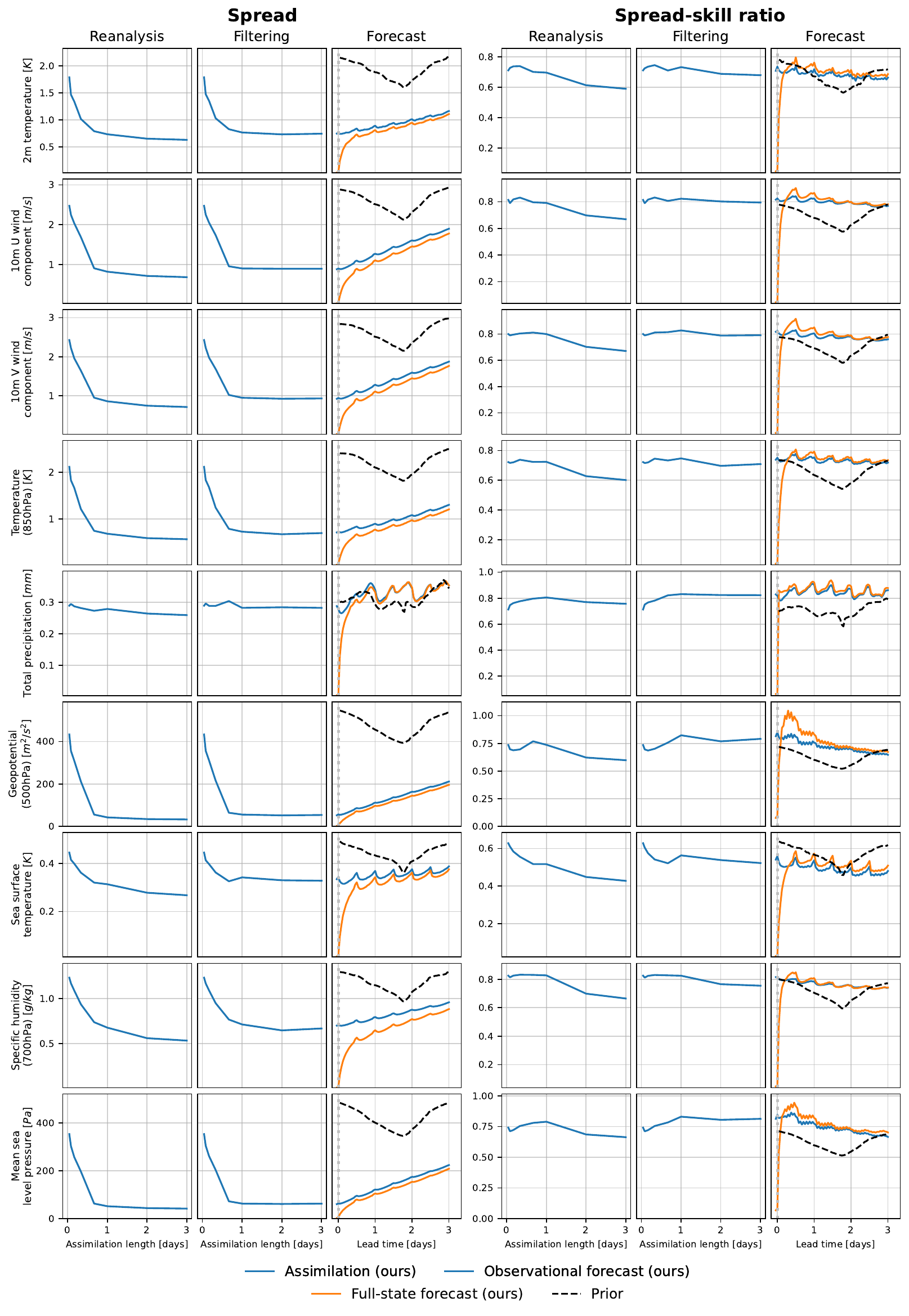}
    \caption{Quantitative evaluation of reanalysis, filtering, and forecasting. (Left) Spread and (Right) Spread-skill ratio for representative variables. Reanalysis scores are averaged over assimilation windows, while filtering reports the last reanalyzed state. Ensemble spread decreases with longer windows, while ratios remain fairly unchanged. A ratio below one indicates overconfidence.}
\end{figure}

\subsection{Qualitative snapshots} \label{app:gallery}

\begin{figure}[H]
    \centering
    \includegraphics[width=\linewidth]{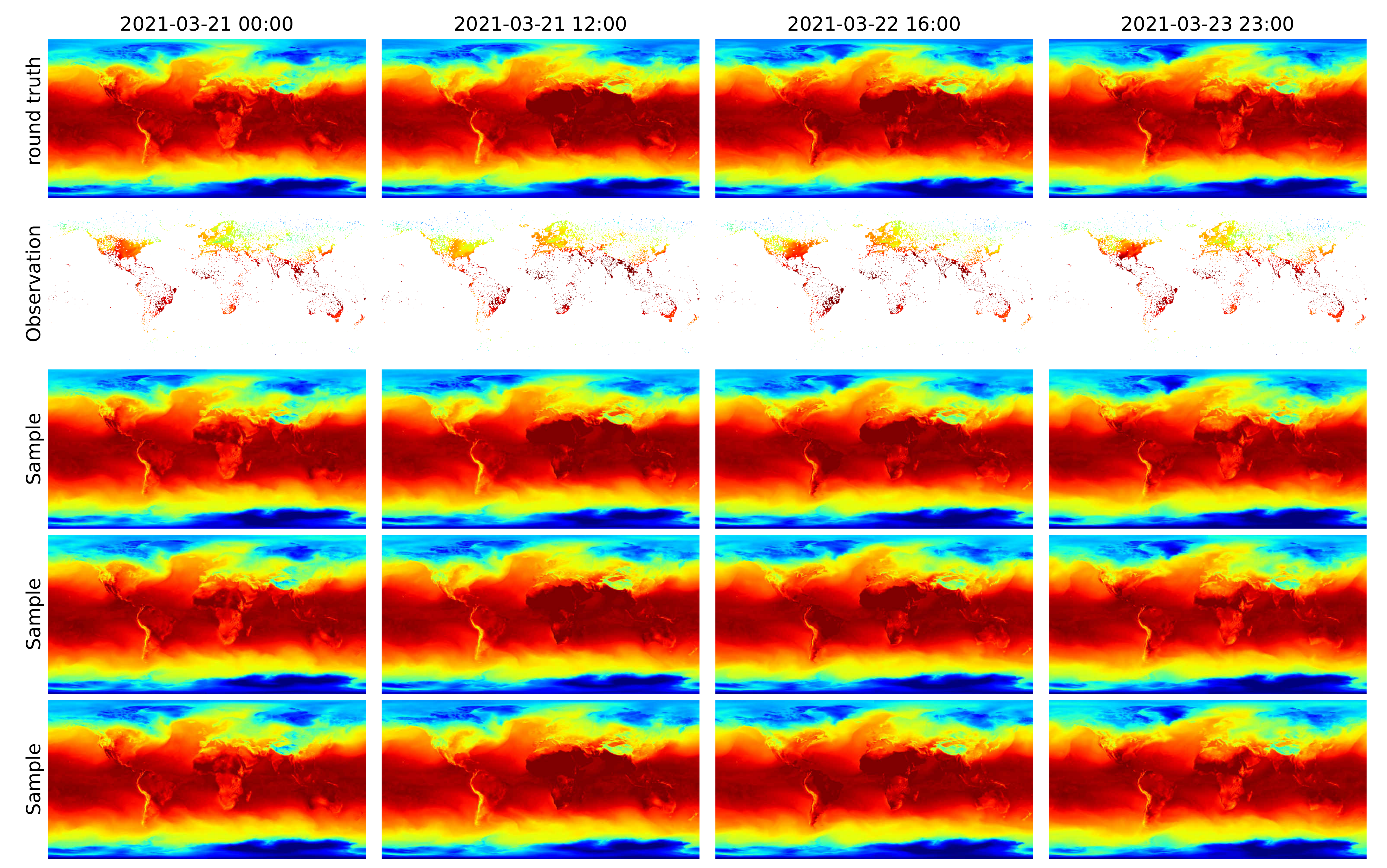}
    \includegraphics[width=\linewidth]{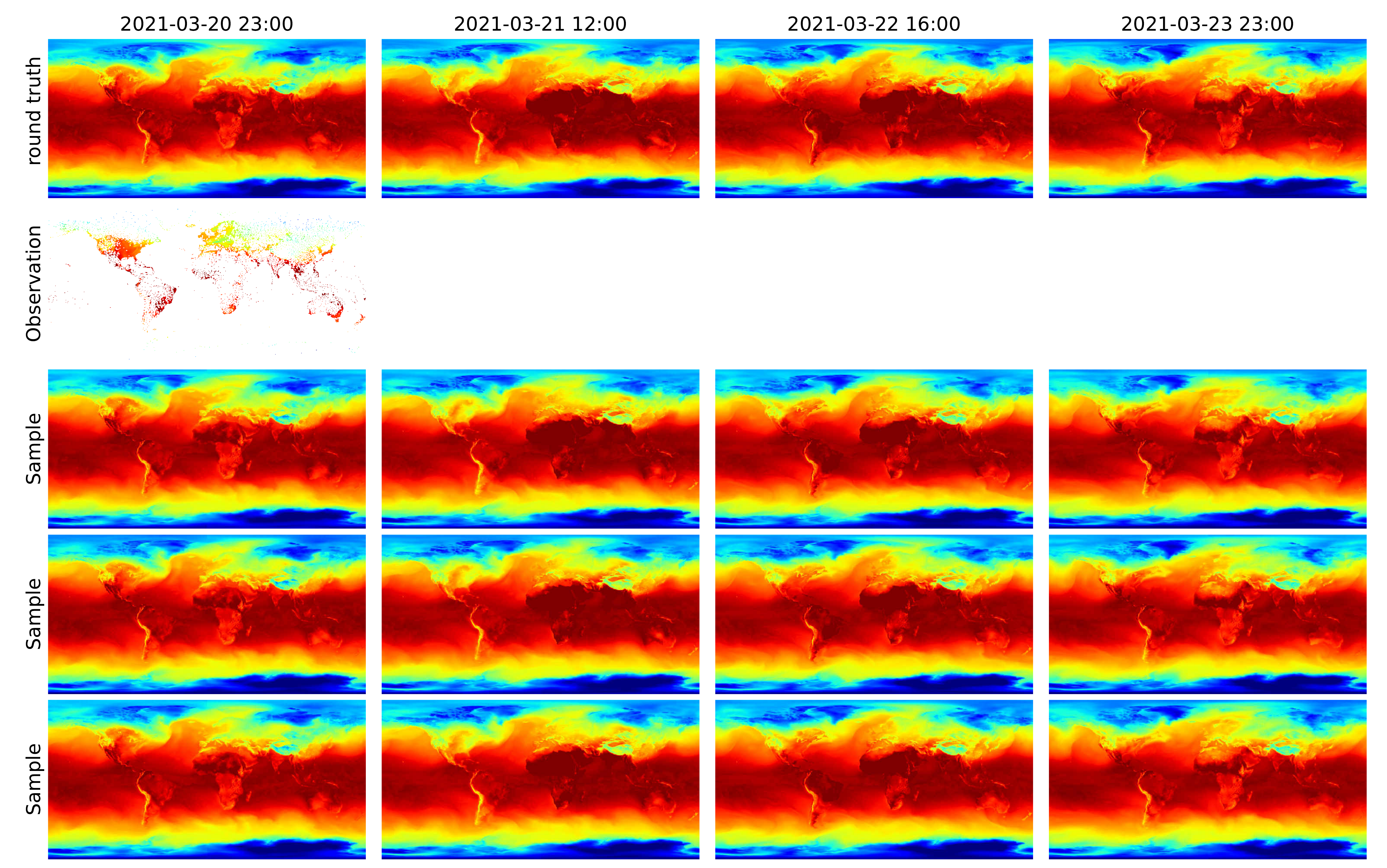}
    \caption{Reconstructed sampled trajectories for surface temperature assimilation. (Top) Reanalysis over a window of 72 hours. (Bottom) Observational forecasting over 3 days initialized with the last 12 states of an assimilation over 24 hours.}
    \label{fig:gallery_t2m}
\end{figure}

\begin{figure}
    \centering
    \includegraphics[width=\linewidth]{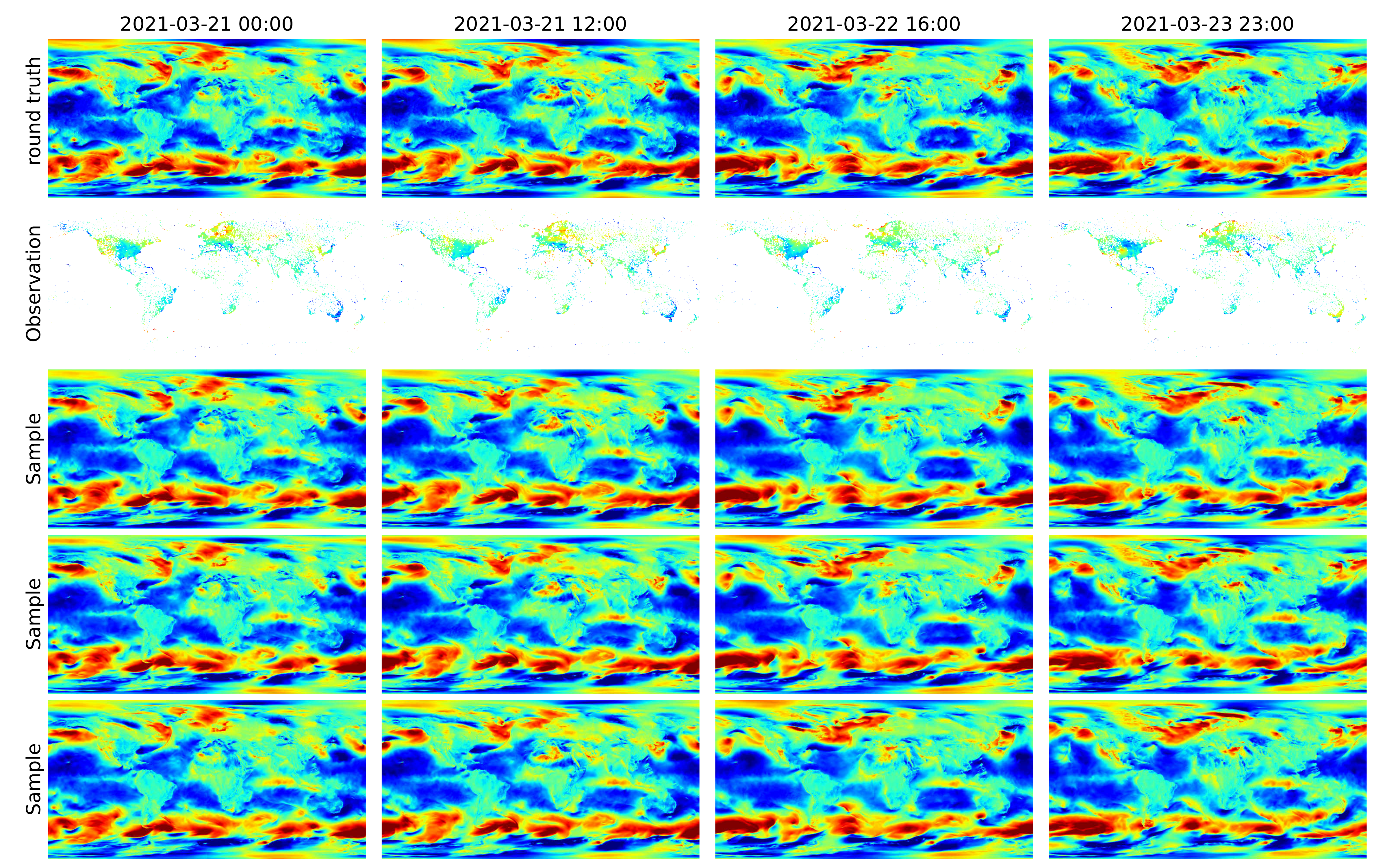}
    \includegraphics[width=\linewidth]{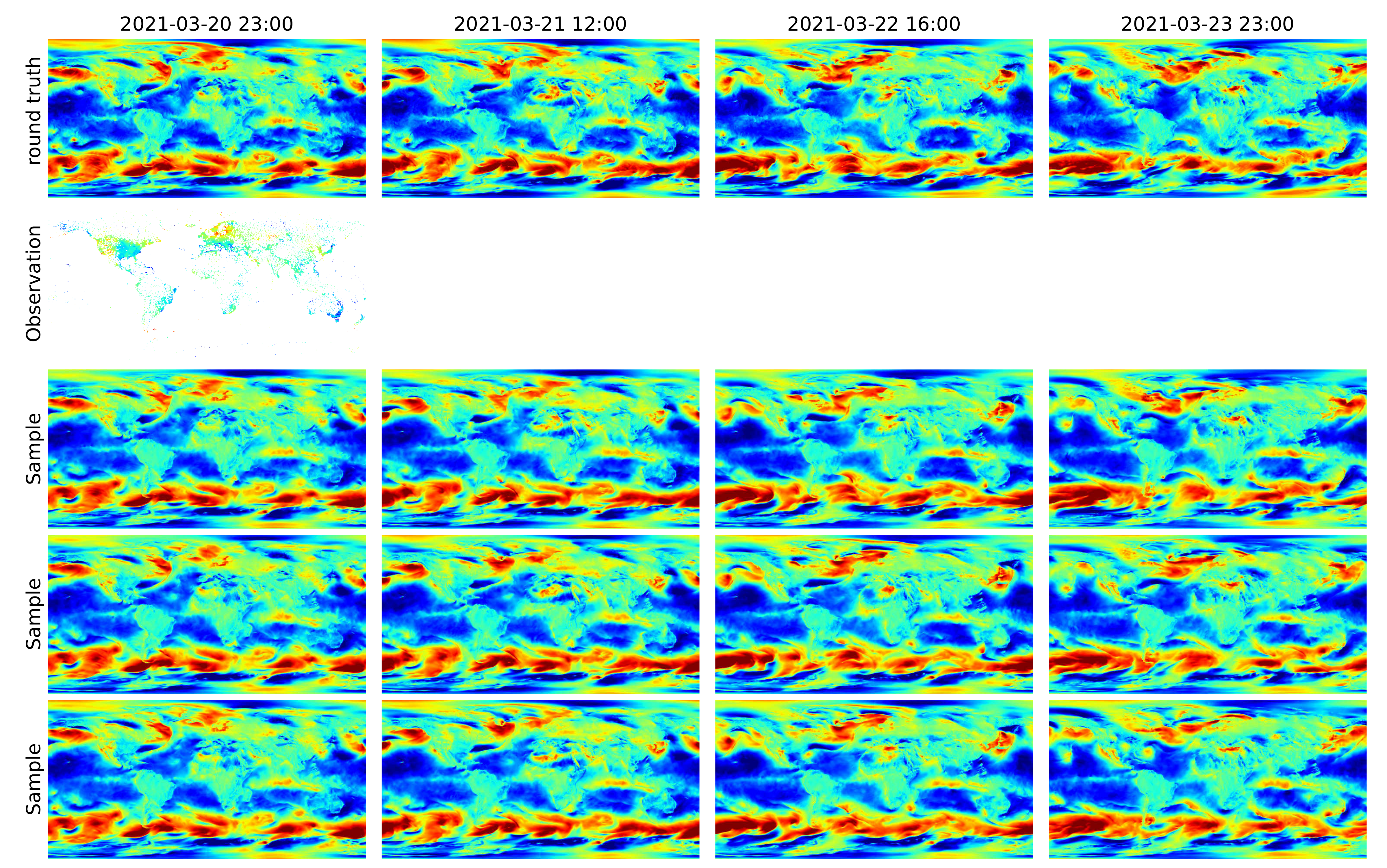}
    \caption{Reconstructed sampled trajectories for surface eastward wind speed assimilation. (Top) Reanalysis over a window of 72 hours. (Bottom) Observational forecasting over 3 days initialized with the last 12 states of an assimilation over 24 hours.}
    \label{fig:gallery_10u}
\end{figure}

\begin{figure}
    \centering
    \includegraphics[width=\linewidth]{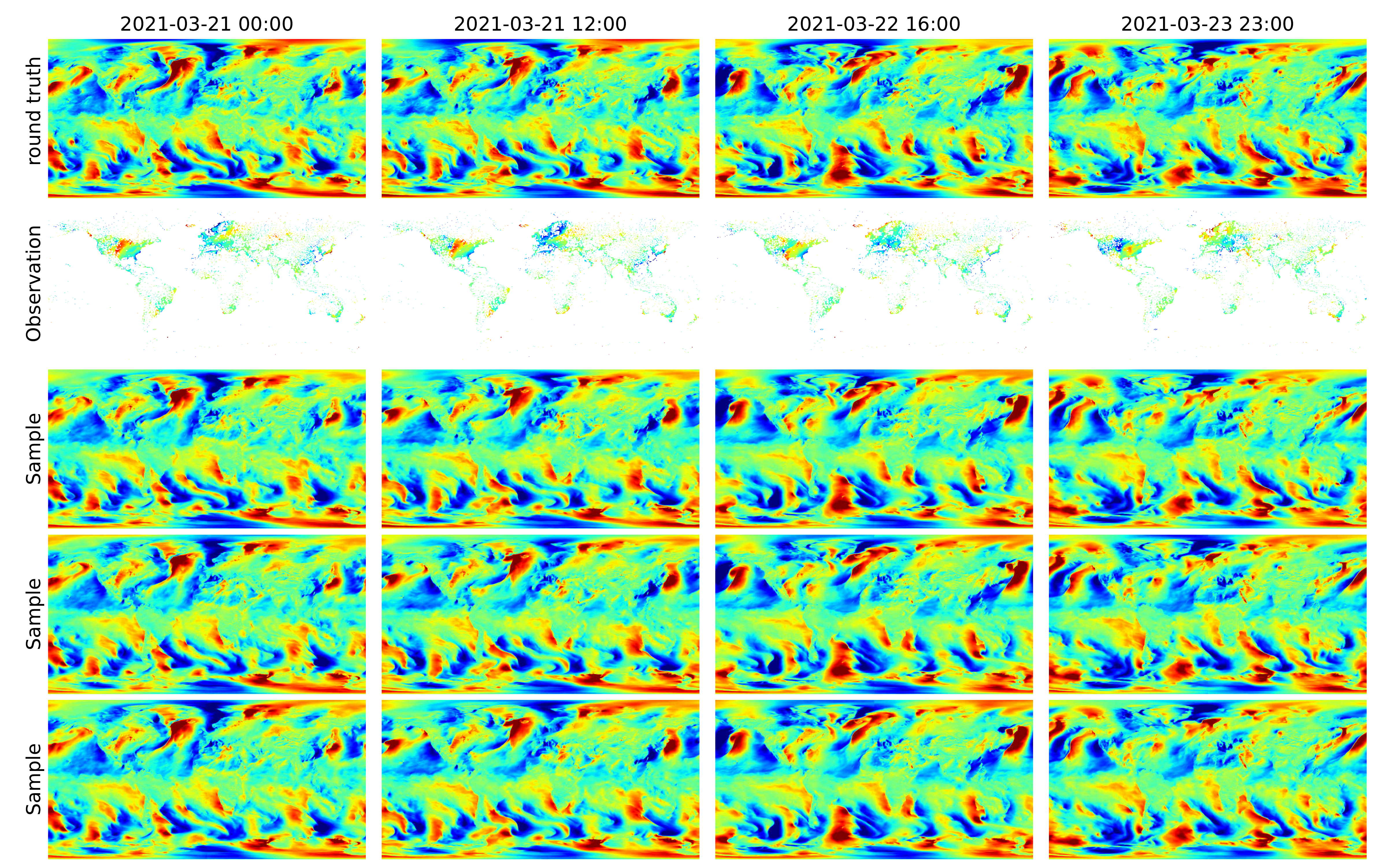}
    \includegraphics[width=\linewidth]{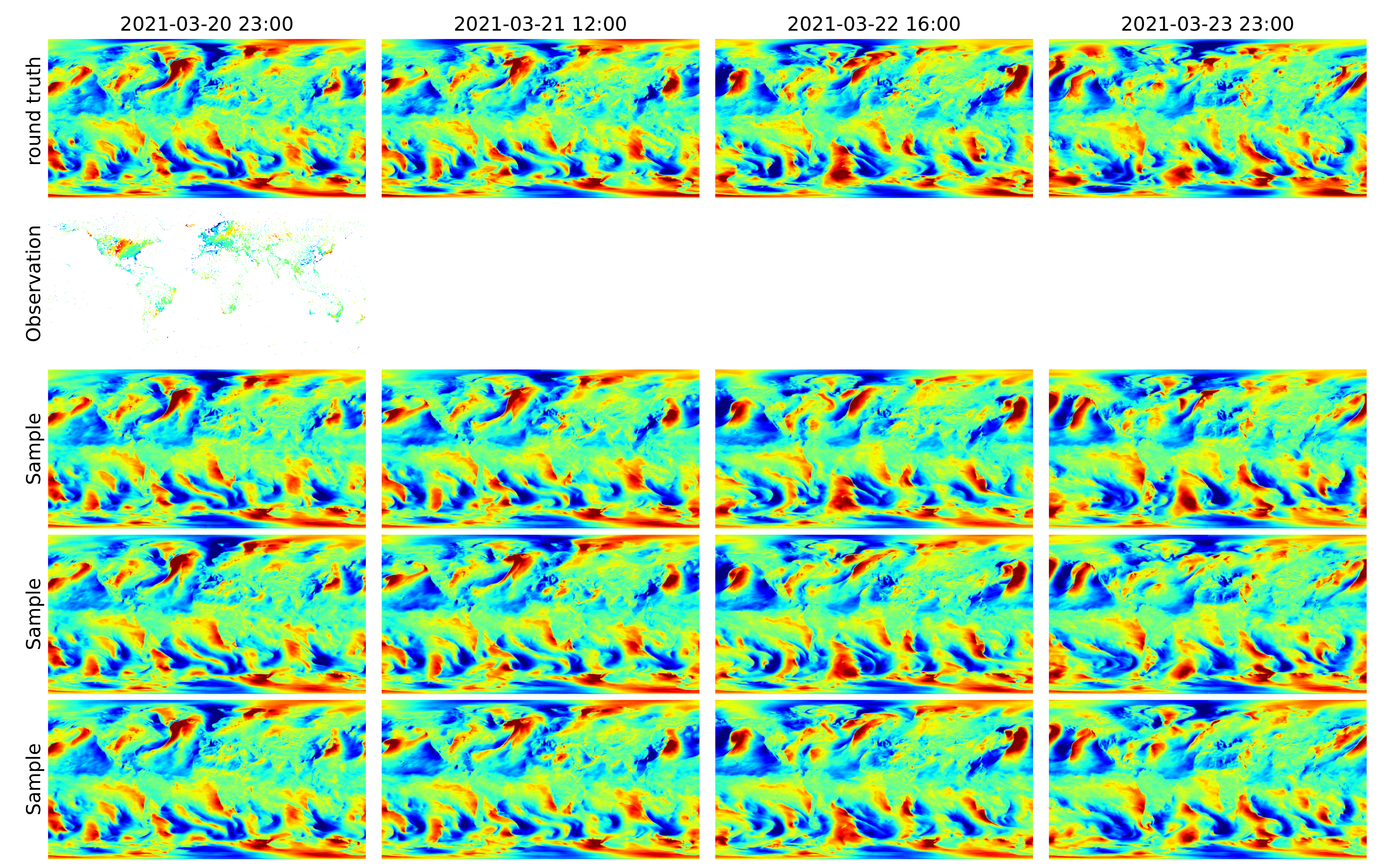}
    \caption{Reconstructed sampled trajectories for surface northward wind speed assimilation. (Top) Reanalysis over a window of 72 hours. (Bottom) Observational forecasting over 3 days initialized with the last 12 states of an assimilation over 24 hours.}
    \label{fig:gallery_10v}
\end{figure}

\begin{figure}
    \centering
    \includegraphics[width=\linewidth]{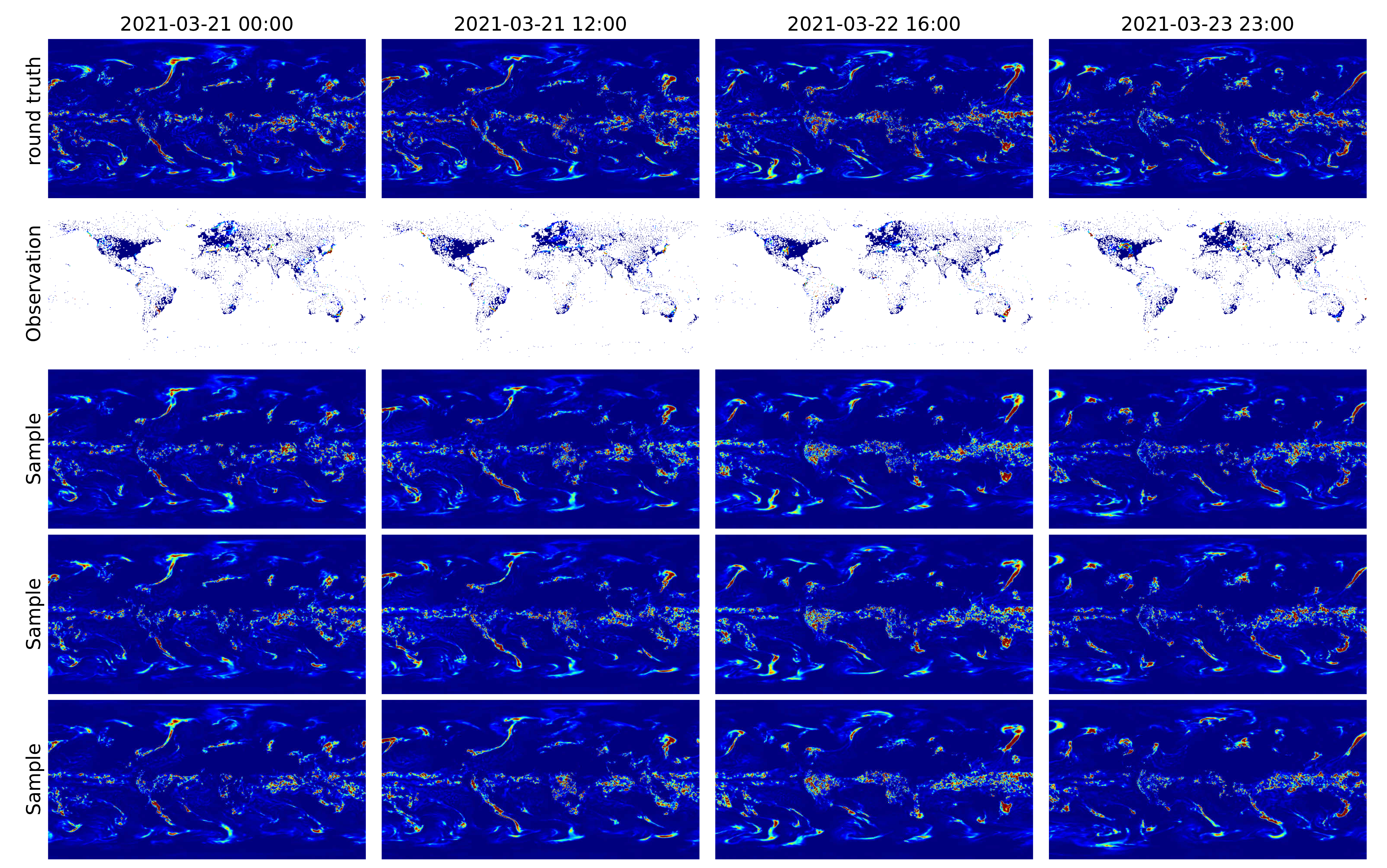}
    \includegraphics[width=\linewidth]{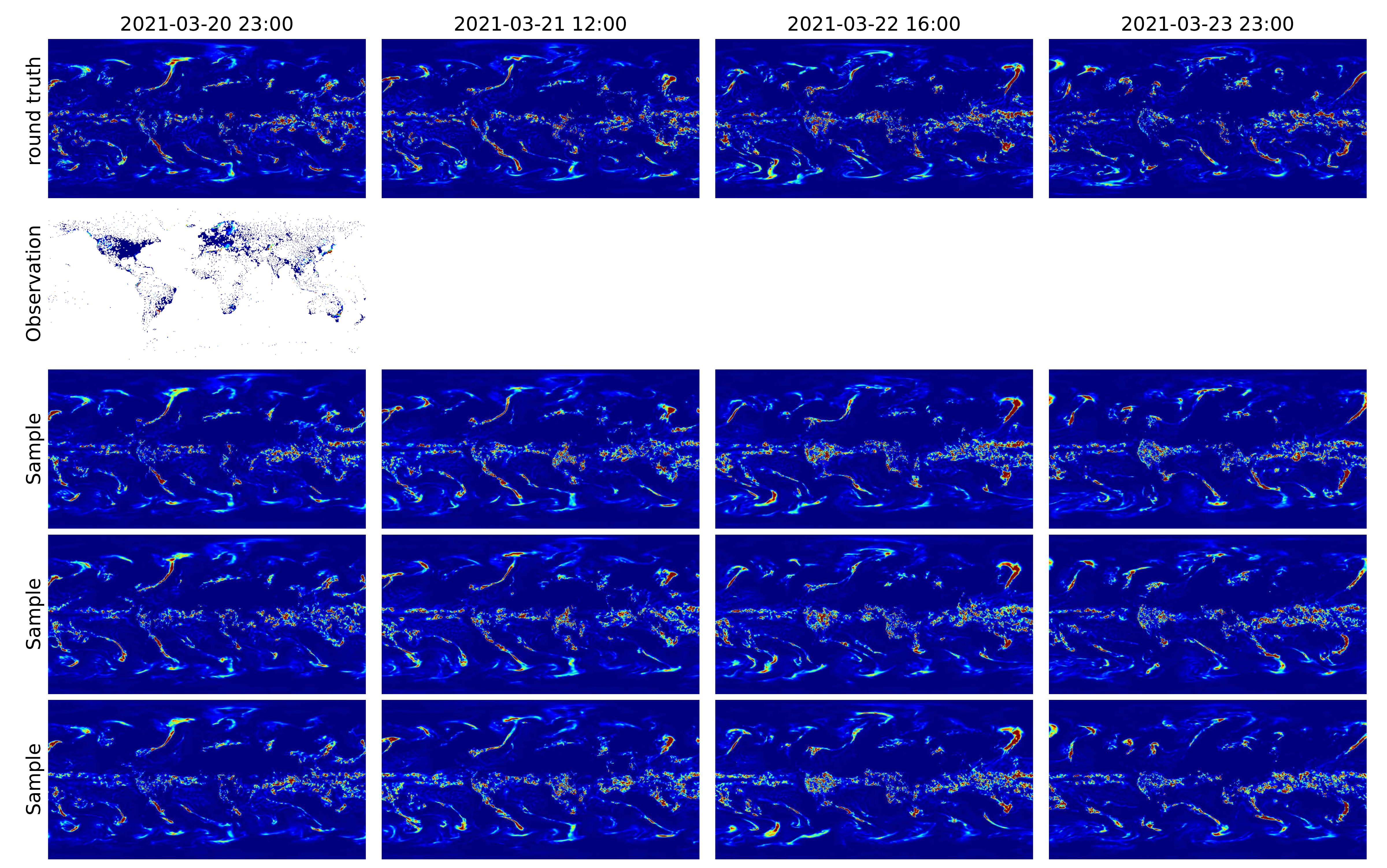}
    \caption{Reconstructed sampled trajectories for total precipitation assimilation. (Top) Reanalysis over a window of 72 hours. (Bottom) Observational forecasting over 3 days initialized with the last 12 states of an assimilation over 24 hours.}
    \label{fig:gallery_tp}
\end{figure}

\begin{figure}
    \centering
    \includegraphics[width=\linewidth]{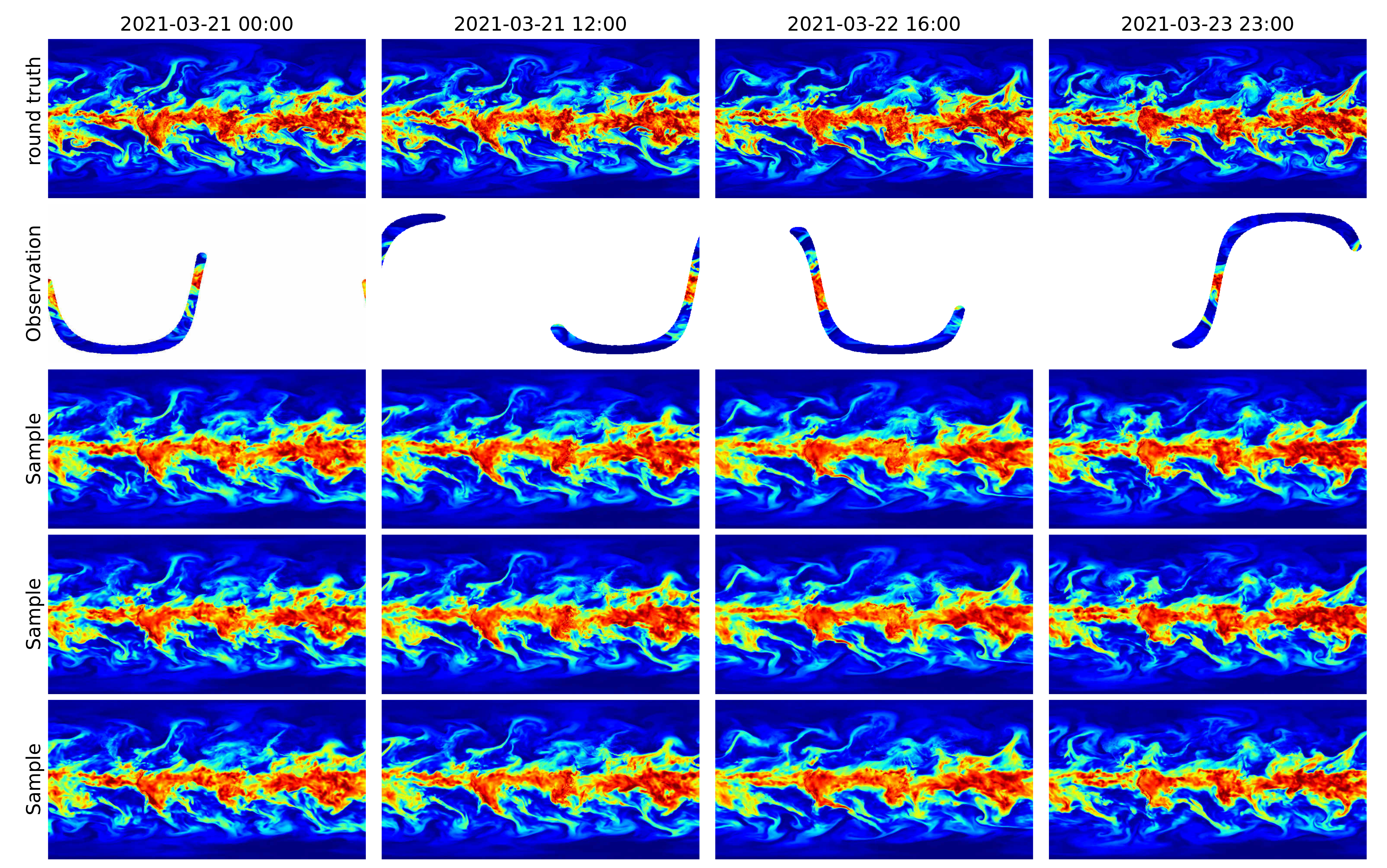}
    \includegraphics[width=\linewidth]{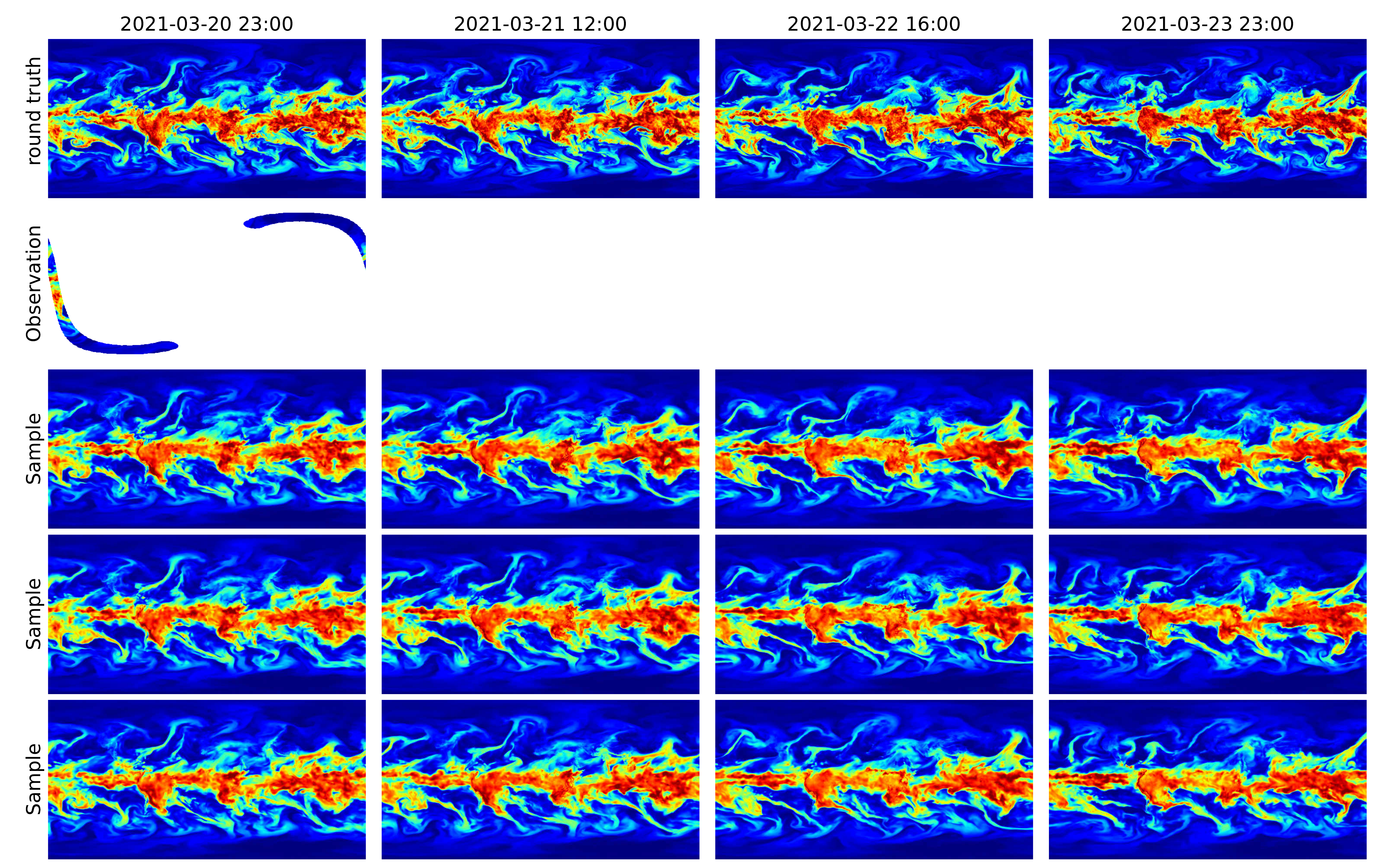}
    \caption{Reconstructed sampled trajectories for specific humidity assimilation at 700hPa. (Top) Reanalysis over a window of 72 hours. (Bottom) Observational forecasting over 3 days initialized with the last 12 states of an assimilation over 24 hours.}
    \label{fig:gallery_q700}
\end{figure}

\begin{figure}
    \centering
    \includegraphics[width=\linewidth]{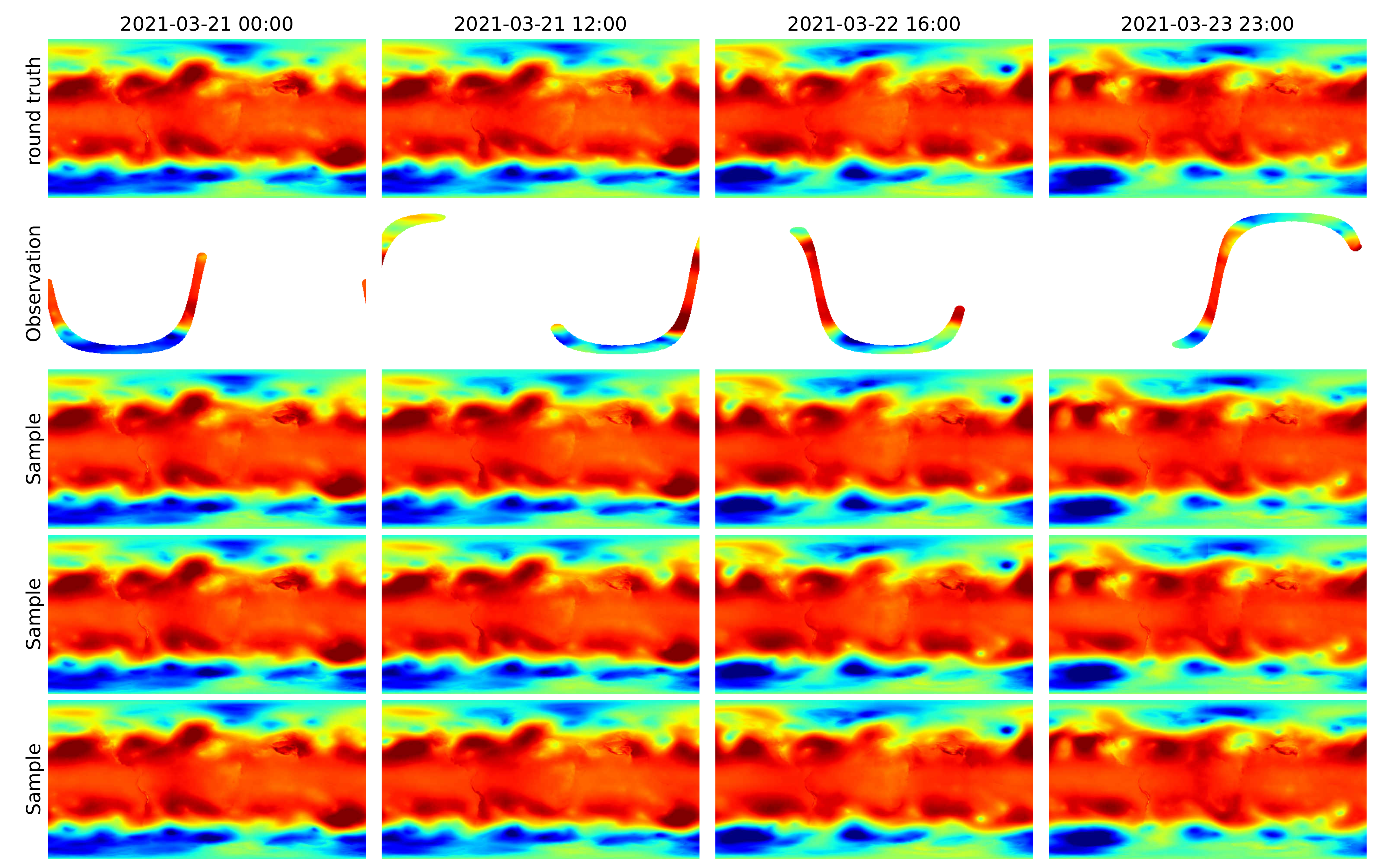}
    \includegraphics[width=\linewidth]{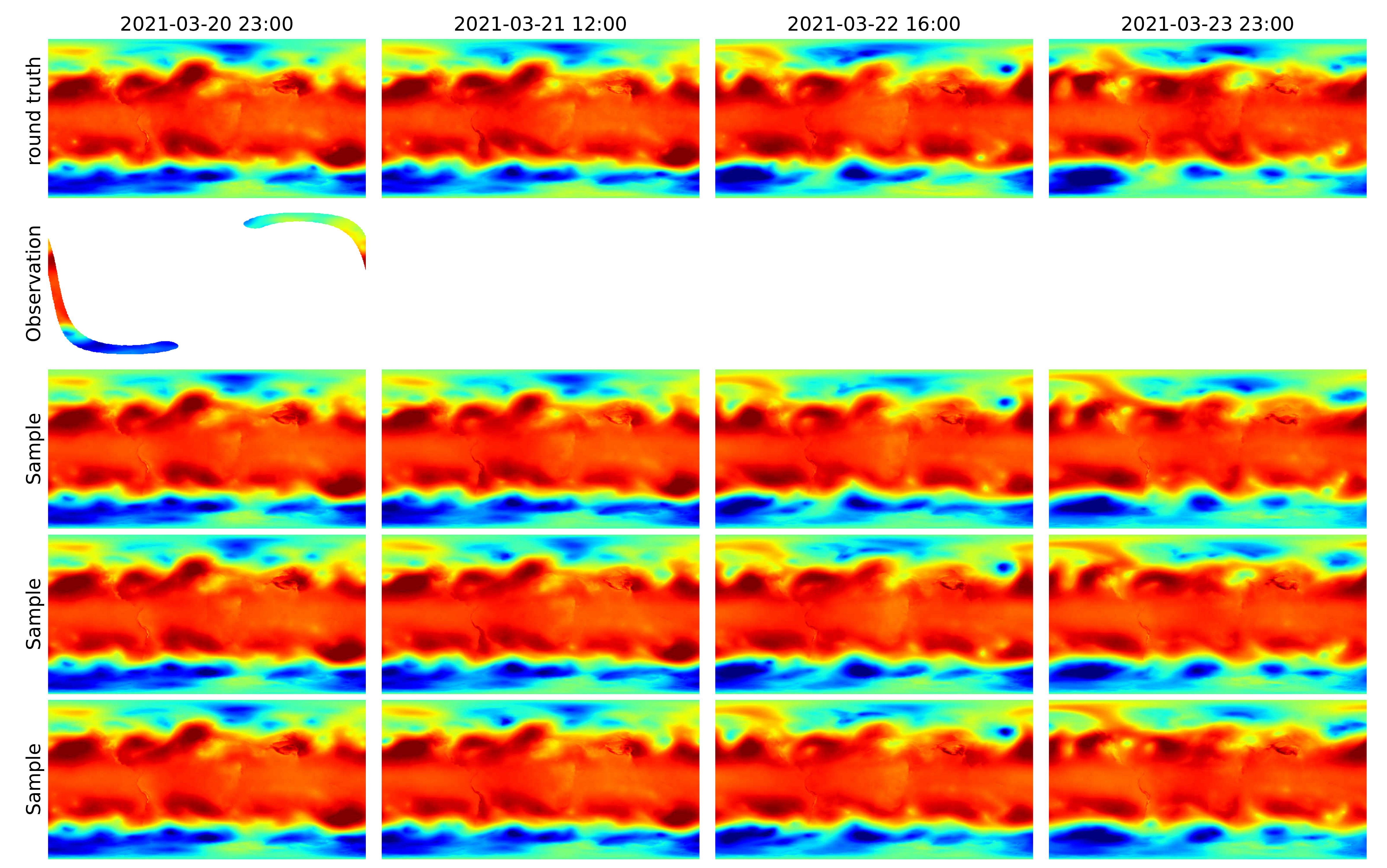}
    \caption{Reconstructed sampled trajectories for geopotential assimilation at 850hPa. (Top) Reanalysis over a window of 72 hours. (Bottom) Observational forecasting over 3 days initialized with the last 12 states of an assimilation over 24 hours.}
    \label{fig:gallery_z850}
\end{figure}

In Figures~\ref{fig:gallery_t2m} to \ref{fig:gallery_z850}, we display decoded sampled trajectories generated through reanalysis over a window of three days as well as through observational forecasting initialized from reanalyzed states, for six representative variables. The second row of each gallery shows the observed pixels, if the state was observed. Each mask is displayed as the corresponding variable was observed during inference, i.e., as ground stations for surface variables and satellite scans for atmospheric variables.

\end{document}